%% file: main.tex
\documentclass[10pt,twocolumn,letterpaper]{article}

\usepackage{cvpr} 

\input{preamble}

\definecolor{cvprblue}{rgb}{0.21,0.49,0.74}
\usepackage[pagebackref,breaklinks,colorlinks,allcolors=cvprblue]{hyperref}

\def\confName{CVPR}
\def\confYear{2027}

\title{\ours{}: Co-Designing Phase-Specialised Speculative Decoding\\and Visual Token Pruning for Efficient GUI Agents}

\author{
Guanxi Lu$^{1,2}$\quad
Young D. Kwon$^{2}$\quad
Miles Williams$^{2}$\quad
Stylianos I. Venieris$^{2}$\\
$^{1}$Imperial College London, UK\qquad
$^{2}$Samsung AI Center, Cambridge, UK\\
{\tt\small guanxi.lu22@imperial.ac.uk, \{yd.kwon, mil.williams, s.venieris\}@samsung.com}
}

\begin{document}
\maketitle
\input{sec/0_abstract}    
\input{sec/1_intro}
\input{sec/2_background}
\input{sec/3_spd}

\input{sec/4_vtp}
\input{sec/5_evaluation}
\input{sec/6_conclusion}

{
    \small
    \bibliographystyle{ieeenat_fullname}
    \bibliography{main}
}

\input{sec/X_suppl}

\end{document}

%% file: preamble.tex
\usepackage{microtype}
\renewcommand{\paragraph}[1]{\vspace{.5em}\noindent\textbf{#1.}}
\usepackage{xspace}
\usepackage{soul}
\usepackage{enumitem}
\usepackage{multirow}
\usepackage{float}
\usepackage[most]{tcolorbox}
\usepackage{pifont}
\usepackage{cuted}

\usepackage{amsmath,amssymb}
\usepackage{algorithm}
\usepackage{algpseudocode}
\algrenewcommand{\algorithmicrequire}{\textbf{Input:}}
\algrenewcommand{\algorithmicensure}{\textbf{Output:}}

\AtBeginDocument{%
  \crefname{figure}{Figure}{Figures}%
  \crefname{table}{Table}{Tables}%
  \crefname{equation}{Equation}{Equations}%
  \crefname{section}{Section}{Sections}%
  \crefname{subsection}{Section}{Sections}%
  \crefname{subsubsection}{Section}{Sections}%
  \crefname{appendix}{Appendix}{Appendices}%
  \crefname{subappendix}{Appendix}{Appendices}%
  \crefname{algocf}{Algorithm}{Algorithms}%
  \crefformat{equation}{#2Equation~#1#3}%
  \Crefformat{equation}{#2Equation~#1#3}%
  \crefmultiformat{equation}{#2Equations~#1#3}{ and~#2#1#3}{, #2#1#3}{ and~#2#1#3}%
  \crefrangeformat{equation}{#3Equations~#1#4 to~#5#2#6}%
}

\newcommand{\ours}{SpecGUI}
\newcommand{\SpdLong}{Phase-Specialised Speculative Decoding}
\newcommand{\SpdShort}{PS-SD}
\newcommand{\LookupLong}{Survival-Guided Hybrid Drafting}
\newcommand{\LookupShort}{SGHD}
\newcommand{\RegexLong}{Schema-Constrained Decoding}
\newcommand{\RegexShort}{SCD}
\newcommand{\VtpLong}{Drafter-Guided Historical Pruning}
\newcommand{\VtpShort}{DGHP}

\newif\ifshownotes
\shownotestrue

\usepackage{tabularx}
\usepackage{booktabs}
\usepackage{array}
\usepackage{multicol}
\usepackage{multirow}
\definecolor{gainGreen}{RGB}{34,139,69}

%% file: sec/0_abstract.tex
\begin{abstract}
Vision-Language Model (VLM)-based GUI agents operate in multi-turn cycles, observing the screen and determining the next action. This process is computationally expensive: accumulating screenshots incurs substantial prefill overhead, while generating reasoning and actions demands extensive decoding. We present SpecGUI, a training-free, plug-and-play framework that co-designs speculative decoding and visual token pruning for efficient GUI agent inference. First, to accelerate decoding, we observe a pronounced asymmetry in the decoding stage. The long, open-ended reasoning phase is hard to predict, whereas the short, structured action phase is easy to draft. To exploit this asymmetry, we propose Phase-Specialised Speculative Decoding, combining Survival-Guided Hybrid Drafting for reasoning with Schema-Constrained Decoding for actions. Second, to reduce the remaining prefill latency, we study the interaction between speculative decoding and visual token pruning. We find that acceptance length is more robust to pruning historical screenshots than the current one, and that drafter attention effectively highlights task-relevant regions. Motivated by this, we propose Drafter-Guided Historical Pruning to guide historical visual token pruning with drafter attention. We implement SpecGUI in SGLang and evaluate it across three GUI agents and two diffusion drafters. Phase-Specialised Speculative Decoding achieves up to $3.35\times$ decoding speedup over autoregressive baselines. Drafter-Guided Historical Pruning prunes 90\% of historical visual tokens while retaining 96.8\% of baseline accuracy on average. Our work sets a new frontier for full-pipeline GUI agent acceleration.
\end{abstract}

%% file: sec/1_intro.tex
\vspace{-5mm}

\section{Introduction}
\label{sec:intro}

\begin{figure*}[t]
    \centering
    \includegraphics[width=0.95\linewidth]{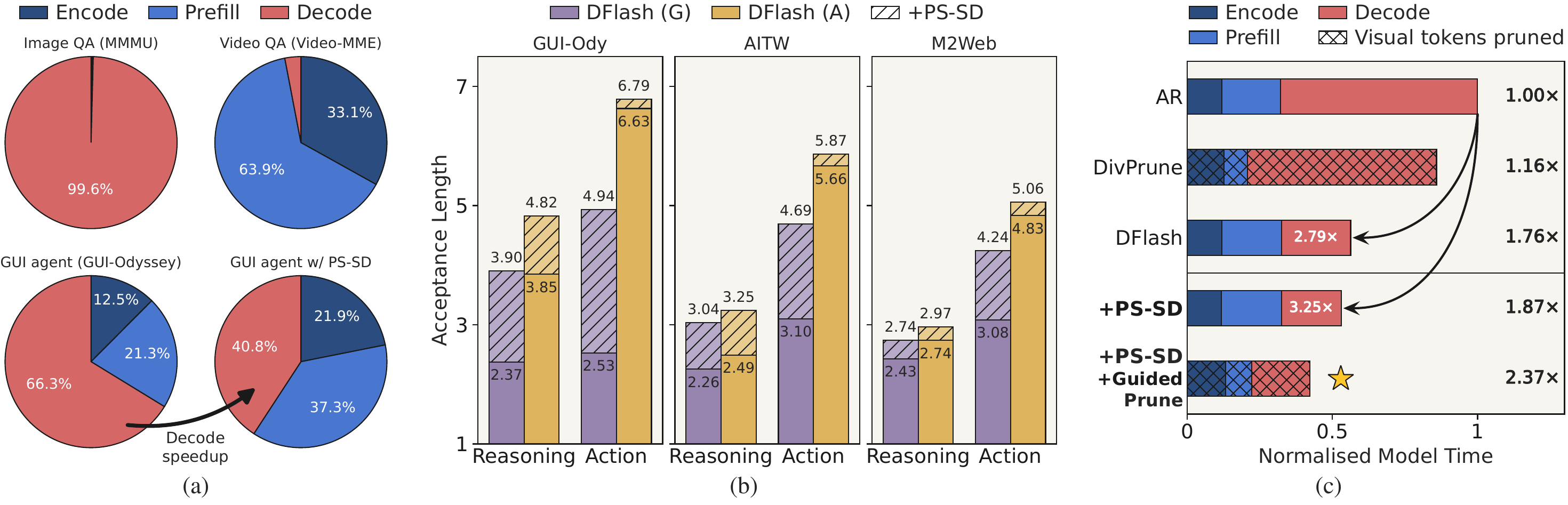}
    \vspace{-3mm}
    \caption{
    \textbf{(a)} Latency breakdown of Qwen3-VL-8B on image QA, video QA, and GUI agent tasks.
    \textbf{(b)} Phase asymmetry: short actions yield high acceptance length, especially after in-domain fine-tuning.
    \textbf{(c)} \ours{} significantly accelerates both the prefill and decoding stages.
    }
    \vspace{-5mm}
    \label{fig:intro}
\end{figure*}

Vision-language models (VLMs) increasingly underpin agents that interact with graphical user interfaces (GUIs). These agents can operate mobile devices, desktops, and web browsers on behalf of users~\cite{cogagent,seeclick,uitars,mobileagentv35,qwenuiagent}. GUI agents function through a multi-turn cycle of observing the screen and performing actions. At each turn, the VLM encodes a screenshot of the current interface into visual tokens, then performs prefill over these tokens and preceding turns. Subsequently, it decodes reasoning tokens to determine the next action, followed by the action itself in a structured format. Finally, the agent parses the action and executes it on the device. Together, these stages take several seconds per turn with 8B models even on an NVIDIA A100 GPU, leaving users waiting minutes on tasks spanning dozens of turns~\cite{rawles2025androidworld}.

For reasoning-enabled GUI agents~\cite{mobileagentv35,maiui}, the primary bottleneck is the autoregressive decoding. Speculative decoding~\cite{leviathan2023fast,chen2023accelerating} seeks to accelerate decoding by cheaply drafting a sequence of tokens and verifying them in parallel. Recent block-diffusion variants~\cite{dflash,dspark} offer further acceleration by proposing a block of candidate tokens in parallel. In GUI agents, decoding exhibits a pronounced asymmetry (\cref{fig:intro}b) between the reasoning and action phases. Specifically, the reasoning phase is long, open-ended, and hard to predict, whereas the action phase is short, structured, and much easier to draft. Applying speculative decoding uniformly across both phases overlooks this asymmetry, resulting in \textit{block inefficiency} and missed \textit{agent-specific opportunities}. First, a fixed block size is computationally inefficient: a large block size wastes compute verifying low-confidence reasoning tokens, while a small one limits the acceptance length achievable for actions. Second, existing methods fail to exploit GUI-specific characteristics, such as cross-turn repetition and schema constraints, to improve acceptance.

Motivated by this, we propose \underline{P}hase-\underline{S}pecialised \underline{S}peculative \underline{D}ecoding (\SpdShort{}, \cref{sec:pssd}), which combines \textit{\LookupLong{}} for the reasoning phase with \textit{\RegexLong{}} for the action phase. \LookupLong{} uses adaptive block allocation to dynamically replace low-confidence block suffixes with candidates retrieved from intra- and inter-episode repetitions. Correspondingly, in the action phase, \RegexLong{} directly populates schema-determined positions and constrains draft tokens to candidates that comply with the schema. Together, these two techniques yield consistent improvements in acceptance length that translate into substantial decoding speedups.

While \SpdShort{} offers substantial acceleration, optimising the decoding stage alone is insufficient for maximal end-to-end speedup. In traditional workloads such as visual question answering~\cite{mmmu,antol2015vqa} and video understanding~\cite{videomme,longvideobench}, latency is overwhelmingly dominated by either decoding or prefill (\cref{fig:intro}a). However, GUI agents face a workload where the latency is more balanced between decoding and prefill. For example, applying \SpdShort{} reduces the decoding share from 66.3\% to 40.8\%, while the prefill share rises from 21.3\% to 37.3\%. This substantial prefill cost stems from high-resolution inputs for accurate coordinate grounding~\cite{zhang2025phi} and historical screenshots accumulated from previous turns~\cite{showui}. 

Visual token pruning~\cite{fastv,divprune} offers a natural approach to reducing prefill overhead, yet its interaction with speculative decoding in GUI agents remains unexplored. Examining this interaction yields two key insights (\cref{sec:vtp}). First, acceptance length is more robust to pruning historical screenshots than the current one. Second, drafter attention serves as an effective signal for identifying task-relevant regions. Leveraging these findings, we propose \textit{\VtpLong{}} to selectively prune historical visual tokens.

We evaluate our approach across three GUI agents~\cite{mobileagentv35,maiui,qwen3vl} and two diffusion drafters~\cite{dflash,domino}. Our results show that~\SpdShort{} consistently improves acceptance length and speedup on both drafters, achieving up to $3.35\times$ decoding speedup over autoregressive baselines. When combined with~\VtpLong{}, our method preserves 96.8\% of baseline accuracy on average with history pruned by 90\%. To reflect production-level performance, we implement our approach in the widely adopted SGLang~\cite{sglang} inference engine. Our work presents a training-free, plug-and-play framework that co-designs speculative decoding and visual token pruning for efficient GUI agent inference.

\noindent
In this paper, we make three key contributions:
\begin{enumerate}[left=0pt,nosep]
    \item We propose \SpdLong{}, to our knowledge the first phase-specialised, training-free speculative decoding framework for GUI agents, exploiting reasoning--action asymmetry with \LookupLong{} and \RegexLong{}.
    \item We present the first systematic study of the interaction between speculative decoding and visual token pruning in GUI agents, and derive practical guidelines for jointly deploying them for end-to-end inference acceleration.
    \item We implement~\ours{} in SGLang and evaluate it across three models, achieving up to a 3.35$\times$ decoding speedup and retaining 96.8\% of baseline accuracy on average when retaining only 10\% of historical visual tokens.
\end{enumerate}

\vspace{-3mm}

%% file: sec/2_background.tex
\section{Background and Related Work}
\label{sec:background}

\subsection{Related Work}
\label{subsec:related}

\paragraph{Efficient GUI Agents}
VLM-based GUI agents~\cite{cogagent,seeclick} build on vision-language models~\cite{flamingo,blip2,llava} to map screen observations and user instructions to executable actions. Recent GUI agents increasingly unify perception, reasoning, and action within a single model trained on GUI data~\cite{showui,uitars,uitars2,mobileagentv35,qwenuiagent,uivenus2,survey_guiagents,survey_benchmarks}. As these agents rely on VLM backbones, general-purpose VLM acceleration techniques can reduce vision-encoding and prefill overhead through efficient vision encoders~\cite{fastvlm,metaxas2026ultravit}, visual token pruning~\cite{visionzip,fastv,pyramiddrop,sparsevlm,divprune,zooprune}, layer skipping~\cite{shortv}, and KV cache reuse~\cite{vlacache,vlcache,kamera}. Recent works further propose efficiency optimisations tailored to GUI agents, including visual token pruning~\cite{ouyang2026focusui,guipruner,aquaui,histprune}, KV cache compression~\cite{guikv,starkv}, context simplification~\cite{simpagent,secagent}, on-device resource utilisation~\cite{pastforward,mobileexplorer,chen2026aospec}, and full-pipeline acceleration~\cite{li2026guiflow}. However, these methods primarily reduce prefill or KV cache costs, while the agent still decodes one token per forward pass. For reasoning-enabled GUI agents, this autoregressive decoding cost accounts for a significant portion of end-to-end latency.

\vspace{-1mm}

\paragraph{Speculative Decoding}
Speculative decoding reduces decoding latency by using a lightweight drafter to propose candidate tokens that the target model verifies in a single forward pass~\cite{leviathan2023fast,chen2023accelerating}. Drafters are typically trained as autoregressive~\cite{li2024eagle,li2024eagle2,eagle3,specvocab}, parallel multi-head~\cite{medusa}, or diffusion-based~\cite{christopher2025speculative,dflash} models, whereas training-free variants~\cite{samdecoding,suffixdecoding,graft,toolspec,agentspec} retrieve candidate tokens from historical context. To accelerate multimodal LLM decoding, recent works have explored multimodal-aware drafter training~\cite{msd,vispec,massv,huang25specvlm} and training-free test-time optimisations~\cite{ji25specvlm,hsd,mmspec,mmspec_diffusion_survey}. Recently, block-diffusion speculative decoding~\cite{dflash,dart,domino,dflash2,whiflash} has emerged as a powerful paradigm that predicts a block of $B$ tokens in a single drafter forward pass. Our method,~\SpdLong, introduces phase-specific optimisations to tailor block-diffusion drafting to GUI agent inference.

\subsection{Preliminaries}
\label{subsec:formulation}

\vspace{-2mm}

\paragraph{GUI Agent Inference}
GUI agents perform a task through an episode of successive interaction turns. To generate the response $\mathbf{y}_j$ for turn $j$, a target VLM $M$ conditions on the task instruction $u$, the current screenshot $I_j$, and the retained history $\mathcal{H}_j$ comprising prior screenshots and outputs. Reasoning-enabled GUI agents~\cite{mobileagentv35,maiui,uitars} also perform chain-of-thought (CoT) reasoning prior to action generation, which can help improve performance on complex tasks, such as in interactive environments~\cite{zhang2026does}. We formulate generation using the target model's output distribution $p_M$:
$$
\label{eq:agent_generation}
\mathbf{y}_j = \left[\mathbf{y}_j^{\textsc{reasoning}},\, \mathbf{y}_j^{\textsc{action}}\right], \qquad
\mathbf{y}_j \sim p_M(\cdot \mid u, \mathcal{H}_j, I_j).
$$

\paragraph{Latency Model}
Each GUI-agent turn comprises visual encoding, language-model prefill, and autoregressive decoding. The per-turn latency can therefore be decomposed as 
$$
\label{eq:turn_latency}
T_{\mathrm{turn}} = T_{\mathrm{enc}} + T_{\mathrm{pre}} + T_{\mathrm{dec}}
$$
While $T_{\mathrm{enc}}$ is governed by image resolution and the number of screenshots, $T_{\mathrm{pre}}$ and $T_{\mathrm{dec}}$ depend primarily on the input context length and the output length, respectively.

Speculative decoding~\cite{leviathan2023fast,chen2023accelerating} can reduce $T_{\mathrm{dec}}$ by using a lightweight drafter $D$ to propose candidate tokens for parallel verification by the target model $M$. The resulting effective per-token latency $L$ and decoding speedup $\eta$ are given by
$$
\label{eq:sd_latency}
L = \frac{T_{\mathrm{draft}} + T_{\mathrm{verify}}}{\tau}, \quad
\eta = \frac{L_{\mathrm{target}}}{L} = \frac{\tau L_{\mathrm{target}}}{T_{\mathrm{draft}} + T_{\mathrm{verify}}},
$$
where $T_{\mathrm{draft}}$ and $T_{\mathrm{verify}}$ denote the per-iteration drafting and verification latencies, $\tau$ denotes the mean number of accepted tokens per iteration, and $L_{\mathrm{target}}$ denotes the target model's average autoregressive per-token latency.

Targeting the prefill stage, visual token pruning~\cite{fastv,divprune} reduces $T_{\mathrm{pre}}$ by compressing the visual token sequence. Given a screenshot $I_j$, the visual encoder $E$, and a retained-token budget $b$, the pruning operator $\mathcal{C}$ yields
$$
\label{eq:visual_pruning}
\mathbf{V}_j = E(I_j), \qquad
\widetilde{\mathbf{V}}_j = \mathcal{C}(\mathbf{V}_j; b), \qquad
|\widetilde{\mathbf{V}}_j| \le b.
$$

%% file: sec/3_spd.tex
\section{Phase-Specialised Speculative Decoding}
\label{sec:pssd}

\vspace{-1mm}

\begin{figure*}[t]
    \centering
    \includegraphics[width=\linewidth]{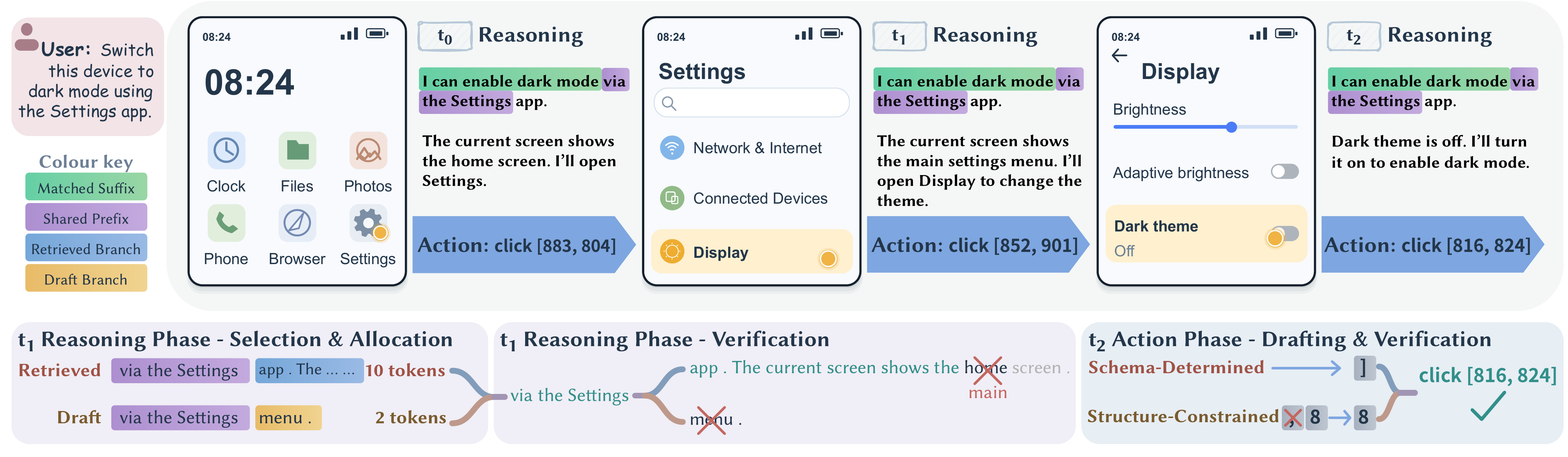}
    \vspace{-7mm}
    \caption{
    \textbf{Illustration of~\SpdLong{}.}
    In a three-turn GUI task,~\LookupLong{} retrieves reasoning tokens from $t_0$ at $t_1$ and improves acceptance length from 4 to 11;~\RegexLong{} revises action tokens at $t_2$.
    }
    \label{fig:method}
    \vspace{-3mm}
\end{figure*}

\subsection{Motivation}
\label{subsec:spd_motivation}

\vspace{-1mm}

The decoding stage of reasoning-enabled GUI agents comprises two asymmetric phases: reasoning and action. The reasoning phase generates longer, free-form sequences to interpret observations and plan decisions. In contrast, the action phase produces concise sequences to serialise tool invocations and their parameters according to a predefined schema. However, applying speculative decoding uniformly across both phases 1) results in \textit{block inefficiency}, and 2) overlooks \textit{agent-specific opportunities} to accelerate decoding. First, under a fixed block size, the longer reasoning phase suffers from a lower acceptance length (\cref{fig:intro}b), causing the target model to waste computation verifying low-confidence tokens. In contrast, setting a smaller block size limits the acceptance length achievable in the action phase. Second, agentic workloads provide phase-specific opportunities to improve acceptance, such as recurring content across past turns and predefined schemas in the action outputs. Although drafters can implicitly learn these common features, their performance heavily depends on whether the training data are exactly in-domain~\cite{dlite}. In comparison, test-time methods offer a more effective and efficient alternative.

Based on these observations, we argue that uniformly applying speculative decoding overlooks phase-specific characteristics, thereby missing optimisation opportunities. We therefore propose \underline{P}hase-\underline{S}pecialised \underline{S}peculative \underline{D}ecoding (\SpdShort), a framework combining \textit{\LookupLong{}} for the reasoning phase with \textit{\RegexLong{}} for the action phase~(\cref{alg:pssd}).

\vspace{-2mm}

\subsection{Survival-Guided Hybrid Drafting (\LookupShort{})}
\label{subsec:hybrid_draft}

During the reasoning phase, low draft acceptance wastes compute verifying  candidate tokens that are likely to be rejected, resulting in block inefficiency. Prior work, such as DSpark~\cite{dspark}, applies adaptive verification to reduce this overhead, but production deployment requires additional padding and scheduler adaptations. Moreover, adaptive verification only converts acceptance to higher speedup, instead of improving draft quality. We address this block inefficiency problem from a complementary perspective: \emph{can we find more promising candidate tokens when the draft tokens are weak?} To this end, we observe two recurring patterns in past reasoning traces. Firstly, \textit{intra-episode repetition}, where the agent restates the task goal, screen description, and plan from the previous turn. Secondly, \textit{inter-episode repetition}, consisting mainly of task-agnostic language patterns such as reasoning boilerplate, generic action descriptions, and standard GUI descriptions. Both repetition patterns can be exploited through $n$-gram retrieval, providing an alternative source of candidate tokens that is distinct from the drafter.

\paragraph{Hybrid Drafting Overview}
Incorporating $n$-gram retrieval into diffusion-based speculative decoding poses two key challenges: 1)~\emph{continuation selection} and 2)~\emph{block allocation}. First, $n$-gram retrieval can return several continuations, while only one can be selected for verification. Second, diffusion-based drafting relies upon a fixed block size $B$, thus we need to determine how to allocate the $B$ positions between draft tokens and retrieved tokens.

\paragraph{Continuation Selection}
To address the challenge of continuation selection, we organise retrieval over two sources: the current episode that includes the prompt, previous turns, and tokens generated so far in the current turn; and the reasoning history from previous episodes. Together, these two sources capture both intra- and inter-episode repetition. We prioritise intra-episode continuations when $n$-gram suffix matches of $n \ge 2$ are found. Otherwise, we query the reasoning history from previous episodes. Within each source, we prioritise longer suffix matches and break ties by recency. Following the retrieval procedure in~\cref{alg:retrieve}, we obtain a continuation $\mathbf r=r_{1:|\mathbf r|}$ with $|\mathbf r|\le B-1$, which complements the draft tokens $\mathbf d$ proposed by $D$.

\begingroup
\makeatletter
\let\pssd@originalruled\fs@ruled
\def\fs@ruled{%
  \pssd@originalruled
  \def\@fs@post{\kern2pt\hrule\kern-15pt\relax}%
}
\makeatother

\input{algorithms/PSSD}
\endgroup

\paragraph{Survival-Guided Block Allocation}
To determine block allocation, we first estimate the cumulative survival probabilities $\widehat S_D(k)$ and $\widehat S_R(k\mid n)$ of the draft branch $\mathbf d$ and retrieved branch $\mathbf r$ up to position $k$ (\cref{alg:allocate}). Here, survival denotes that all tokens through position $k$ are accepted by the target model. We then partition the block of size $B$ to maximise the expected acceptance length. Excluding the initial anchor token and the $b_{\mathrm{shared}}$ prefix tokens shared by both branches leaves $K=B-1-b_{\mathrm{shared}}$ positions to allocate. Maximising the expected acceptance length over the retrieved-branch allocation $0\le b_R\le\min\{K,|\mathbf r|-b_{\mathrm{shared}}\}$ yields the optimal branch allocations $b_R^*$ and $b_D^*$:
\begin{equation}
\label{eq:hybrid_allocation}
\begin{aligned}
b_R^*
&= \operatorname*{arg\,max}_{b_R}
\bigg[
\sum_{k=1}^{b_R}\widehat S_R(k\mid n)
+
\sum_{k=1}^{K-b_R}\widehat S_D(k)
\bigg], \\
b_D^*
&= K-b_R^*.
\end{aligned}
\end{equation}
For the draft branch, cumulative survival is estimated by multiplying the drafter's token-level confidence scores $c_i = p_D(d_{b_{\mathrm{shared}}+i})$, where $i \in \{1, \dots, k\}$ indexes the $i$-th branch position after the shared prefix. In contrast, the retrieved branch lacks native confidence scores. We therefore estimate its stepwise survival probability from historical acceptance statistics conditioned on the matched suffix length $n$ and branch position $i$ (\cref{alg:update_acceptance}). Specifically, we track acceptance counts $H_{n,i}$ and verification counts $N_{n,i}$, updating statistics at branch position $i$ only when all preceding tokens are accepted. To regularise sparse early-stage observations, we apply Bayesian smoothing with a prior $p_0=0.5$ of weight $w=1$. The cumulative survival probabilities of the two branches are then given by
\begin{equation}
\label{eq:hybrid_survival}
\widehat S_D(k) = \prod_{i=1}^{k} c_i,
\qquad
\widehat S_R(k\mid n) = \prod_{i=1}^{k} \frac{H_{n,i}+w p_0}{N_{n,i}+w}.
\end{equation}
 
The resulting candidate tree $\mathcal T$ is verified by the target model $M$ in a single forward pass, producing accepted tokens $\mathbf a$ that are appended to $\mathbf y$ and made available for subsequent retrieval (lines 13--16 of~\cref{alg:pssd}).

\subsection{Schema-Constrained Decoding (\RegexShort{})}
\label{subsec:regex}

During the action phase, GUI agents emit structured actions following predefined schemas. The actions are parsed into execution commands such as on-screen GUI operations or external API requests. Although these structured outputs can be learned by training on in-domain data, test-time token revision improves acceptance at negligible cost. In particular, we identify two opportunities: \textit{1)} schema-determined positions require no prediction from the drafter, and \textit{2)} at structure-constrained positions such as coordinate digits, invalid draft tokens can be replaced with valid candidates. We therefore propose \RegexLong{} (lines 8--12 of~\cref{alg:pssd}) to exploit these two opportunities.

Since revising a draft token can alter subsequent schema constraints, \RegexLong{} sequentially traverses the schema automaton $(\delta, q_0, F)$ starting from the decoded action-prefix state $q'=\delta^*(q_0,\mathbf y_{\mathrm{act}})$. At each draft position $i$, the schema matcher determines
\begin{equation}
\label{eq:schema_candidates}
\mathcal C(q')=
\begin{cases}
\{c(q')\}, & \text{schema-determined},\\
\{v:\delta(q',v)\neq q_\bot\}, & \text{otherwise}
\end{cases}
\end{equation}
where $c(q')$ is the next canonical schema token and $q_\bot$ denotes a schema violation. We select the highest-logit schema-valid token $d_i\in C(q')$ and advance $q'\gets\delta(q',d_i)$. Revising terminates upon reaching an action-complete or reject state in $F$. The revision only applies to draft tokens, while the verification remains lossless. 

\vspace{-1mm}

%% file: algorithms/PSSD.tex
{
\begin{algorithm}[t]
\algrenewcommand{\algorithmicindent}{1em}
\caption{Phase-Specialised Speculative Decoding}
\label{alg:pssd}
\algtext*{EndIf}\algtext*{EndFor}\algtext*{EndWhile}\algtext*{EndFunction}
\begin{algorithmic}[1]
\Statex \textbf{Input:} Target model $M$; drafter $D$; multimodal prompt $\mathbf x$; block size $B$; schema automaton $(\delta,q_0,F)$.
\Statex \textbf{Output:} Completed output $\mathbf y$.
\Function{Decode}{$M,D,\mathbf x,B,(\delta,q_0,F)$}
  \State $y_0\sim p_M(\cdot\mid\mathbf x)$; $\mathbf y\gets[y_0]$ \Comment{Target prefill}
  \While{generation is incomplete}
    \State $(\mathbf d,\mathbf z)\gets D(\mathbf y;B)$ \Comment{Draft tokens and logits}
    \If{$\phi(\mathbf y)=\textsc{reasoning}$} \Comment{\LookupShort{}}
      \State $(\mathbf r,n)\gets\Call{Retrieve}{\mathbf y,B}$ 
      \State $(\mathcal T,b_{\mathrm{shared}},b_R^*)\gets\Call{Allocate}{\mathbf d,\mathbf z,\mathbf r,n,B}$
    \Else \Comment{\RegexShort{}}
      \State $q'\gets\delta^*(q_0,\mathbf y_{\mathrm{act}})$
      \For{$i=1,\ldots,B-1$ \textbf{while} $q'\notin F$}
        \State $d_i\gets\operatorname*{arg\,max}_{v\in\mathcal C(q')}z_{i,v}$; $q'\gets\delta(q',d_i)$
      \EndFor
      \State $\mathcal T\gets\mathbf d$
    \EndIf
    \State $\mathbf a\gets\Call{Verify}{M,\mathbf y,\mathcal T}$
    \If{$\phi(\mathbf y)=\textsc{reasoning}$}
      \State $\Call{UpdateAcceptance}{\mathbf a,\mathbf r,n,b_{\mathrm{shared}},b_R^*}$
    \EndIf
    \State $\mathbf y\gets\mathbf y\Vert\mathbf a$
  \EndWhile
  \State \Return $\mathbf y$
\EndFunction
\end{algorithmic}
\end{algorithm}
}

%% file: sec/4_vtp.tex
\section{Co-Designing Speculative Decoding and Visual Token Pruning}
\label{sec:vtp}

Speculative decoding primarily accelerates the decoding stage, leaving visual input encoding and language-model prefill as substantial sources of per-turn latency. A commonly adopted approach to reducing prefill latency is visual token pruning~\cite{fastv,divprune}, which shortens the sequence of visual tokens that are processed by the language model. Combining both techniques targets prefill and decoding latency simultaneously, but visual pruning can affect speculative acceptance by altering the context available to the drafter. Therefore, we systematically examine how visual token pruning interacts with speculative decoding to guide our co-design.

\begin{figure}[t]
    \centering
    \includegraphics[width=\columnwidth]{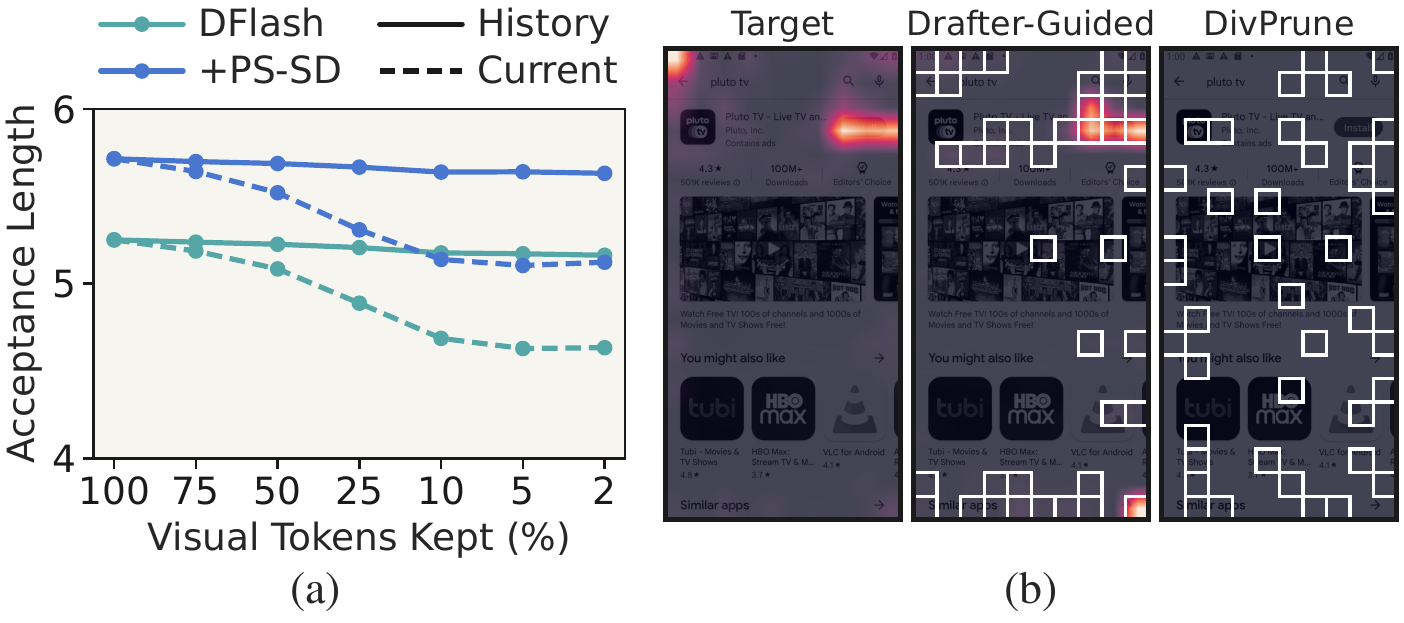}
    \vspace{-6mm}
    \caption{\textbf{(a)} Acceptance length under random pruning for GUI-Owl on AndroidControl. \textbf{(b)} Drafter attention closely aligns with the target model's attention; pre-LLM pruning lacks task awareness.}
    \vspace{-3mm}
    \label{fig:vtp}
\end{figure}

\paragraph{Acceptance is sensitive to pruning the current screenshot, but not historical screenshots}
We observe that pruning historical and current screenshots has asymmetric effects on acceptance (\cref{fig:vtp}a). Pruning historical screenshots has little effect on acceptance length, degrading it by no more than 1.7\% across all tested retention ratios (2\% to 100\%). In contrast, pruning the current screenshot results in a substantial decrease in acceptance length of up to 11.8\%.

This degradation affects both phases, but is more pronounced on tokens that represent on-screen coordinates within the action phase. In GUI agents, historical screenshots provide context about the task progress, whereas the current screenshot supplies fine-grained spatial details for the immediate action. Aggressively pruning the current screenshot removes these details, lowering task accuracy and weakening alignment between the drafter and target model. Moreover, the lightweight drafter mispredicts spatial arguments even when the target model remains accurate. This loss cannot be recovered by~\SpdShort{}, as neither~\LookupShort{} nor~\RegexShort{} restores the visual information removed by pruning.

\input{tables/spd_main}

\paragraph{Drafter-Guided Historical Pruning}
We find that the drafter's attention mechanism can effectively guide the pruning of historical visual tokens. For each historical screenshot, the attention map from its corresponding interaction turn highlights the regions that are most relevant to task progress. Existing \emph{pre}-LLM visual token pruning methods, which prune tokens before they reach the LLM~\cite{divprune,zooprune,cdpruner}, miss this signal because they treat historical and current screenshots identically. In contrast, \emph{in}-LLM methods prune tokens within the LLM layers~\cite{fastv,sparsevlm}. However, these methods compromise efficiency because they are incompatible with the prefix caching used in production inference engines~\cite{sglang,vllm}. We observe that the drafter's attention serves as an effective proxy for the target model's attention (\cref{fig:vtp}b) and can be computed with negligible overhead via a single forward pass of the drafter. Therefore, drafter-guided historical pruning enables efficient pre-LLM token removal while retaining the task awareness of in-LLM visual token pruning methods.

At each turn, we record the drafter's attention map over the current screenshot at the coordinate tokens, or action-type tokens when coordinates are absent. In subsequent turns, historical screenshots are pruned to an overall history retention ratio $b$. To prioritise recent history~\cite{guipruner}, we assign the $t$-th of $T$ chronologically ordered historical screenshots a temporal weight $w_t$ and compute its retention ratio $b_t$ as
\begin{equation}
\label{eq:temporal_weight}
    b_t = b \cdot T \frac{w_t}{\sum_{s=1}^{T} w_s}, \quad w_t = \gamma + (1 - \gamma) \frac{t - 1}{T - 1},
\end{equation}
for $T > 1$ with $\gamma \in (0, 1]$, and set $b_1 = b$ when $T = 1$. The top-$b_t$ fraction of visual tokens with the highest attention scores are retained for each screenshot $t$. The complete algorithm is detailed in~\cref{alg:vtp}.

\vspace{-2mm}

%% file: tables/spd_main.tex
\begin{table*}[t]
\centering
\small
\setlength{\tabcolsep}{0pt}
\renewcommand{\arraystretch}{0.95}
\setlength{\aboverulesep}{0.5ex}
\setlength{\belowrulesep}{0.5ex}
\caption{
{\textbf{Acceptance length and speedup of speculative decoding.}
We report mean acceptance length ($\tau$) and decoding speedup across three target models using general-purpose (G) and GUI-adapted (A) drafters, with autoregressive decoding time shown in ms/turn. GUI-Ody, AITW, and M2Web denote GUI-Odyssey, Android in the Wild, and Mind2Web, respectively. The better result in each pair is in \textbf{bold}.
}
}
\label{tab:spd}
\vspace{-2mm}
\begin{tabular*}{\linewidth}{
    @{\extracolsep{\fill}}
    l
    @{\hspace{0.2em}}
    *{8}{
        c
        @{\extracolsep{0pt}}
        c
        @{\extracolsep{\fill}\hspace{0.2em}}
    }
    c
    @{\extracolsep{0pt}}
    c
    @{}
}
\toprule
& \multicolumn{6}{c}{\textbf{GUI-Owl-1.5-8B}}
& \multicolumn{6}{c}{\textbf{MAI-UI-8B}}
& \multicolumn{6}{c}{\textbf{Qwen3-VL-8B}} \\
\cmidrule(lr){2-7}
\cmidrule(lr){8-13}
\cmidrule(lr){14-19}
& \multicolumn{2}{c}{GUI-Ody}
& \multicolumn{2}{c}{AITW}
& \multicolumn{2}{c}{M2Web}
& \multicolumn{2}{c}{GUI-Ody}
& \multicolumn{2}{c}{AITW}
& \multicolumn{2}{c}{M2Web}
& \multicolumn{2}{c}{GUI-Ody}
& \multicolumn{2}{c}{AITW}
& \multicolumn{2}{c}{M2Web} \\[-0.3ex]
\cmidrule(lr){2-3}
\cmidrule(lr){4-5}
\cmidrule(lr){6-7}
\cmidrule(lr){8-9}
\cmidrule(lr){10-11}
\cmidrule(lr){12-13}
\cmidrule(lr){14-15}
\cmidrule(lr){16-17}
\cmidrule(lr){18-19}
\textbf{Method}
& Speedup & $\tau$ & Speedup & $\tau$ & Speedup & $\tau$
& Speedup & $\tau$ & Speedup & $\tau$ & Speedup & $\tau$
& Speedup & $\tau$ & Speedup & $\tau$ & Speedup & $\tau$ \\
\midrule
AR
& \multicolumn{2}{c}{2133\,ms}
& \multicolumn{2}{c}{1233\,ms}
& \multicolumn{2}{c}{1911\,ms}
& \multicolumn{2}{c}{1006\,ms}
& \multicolumn{2}{c}{916\,ms}
& \multicolumn{2}{c}{1025\,ms}
& \multicolumn{2}{c}{2114\,ms}
& \multicolumn{2}{c}{1822\,ms}
& \multicolumn{2}{c}{3567\,ms} \\
\midrule
DFlash (G)
& 1.66$\times$ & 2.41
& 1.75$\times$ & 2.52
& 1.80$\times$ & 2.59
& 1.31$\times$ & 1.89
& 1.66$\times$ & 2.38
& 1.69$\times$ & 2.41
& 1.40$\times$ & 2.07
& 1.84$\times$ & 2.69
& 1.68$\times$ & 2.42 \\
\quad +\textit{\SpdShort}
& \textbf{2.63}$\times$ & \textbf{4.07}
& \textbf{2.29}$\times$ & \textbf{3.45}
& \textbf{2.08}$\times$ & \textbf{3.08}
& \textbf{2.16}$\times$ & \textbf{3.50}
& \textbf{2.49}$\times$ & \textbf{3.89}
& \textbf{2.30}$\times$ & \textbf{3.46}
& \textbf{2.04}$\times$ & \textbf{3.14}
& \textbf{2.23}$\times$ & \textbf{3.40}
& \textbf{1.93}$\times$ & \textbf{2.82} \\
\midrule
Domino (G)
& 1.63$\times$ & 2.49
& 1.72$\times$ & 2.65
& 1.81$\times$ & 2.77
& 1.21$\times$ & 1.83
& 1.63$\times$ & 2.47
& 1.69$\times$ & 2.55
& 1.60$\times$ & 2.47
& 1.92$\times$ & 2.95
& 1.73$\times$ & 2.62 \\
\quad +\textit{\SpdShort}
& \textbf{2.57}$\times$ & \textbf{4.17}
& \textbf{2.26}$\times$ & \textbf{3.61}
& \textbf{2.07}$\times$ & \textbf{3.25}
& \textbf{2.03}$\times$ & \textbf{3.44}
& \textbf{2.43}$\times$ & \textbf{3.97}
& \textbf{2.23}$\times$ & \textbf{3.57}
& \textbf{2.15}$\times$ & \textbf{3.47}
& \textbf{2.30}$\times$ & \textbf{3.68}
& \textbf{1.96}$\times$ & \textbf{3.04} \\
\midrule[\dimexpr(\lightrulewidth+\heavyrulewidth)/2\relax]
DFlash (A)
& 2.79$\times$ & 4.25
& 2.16$\times$ & 3.15
& 2.13$\times$ & 3.16
& 2.59$\times$ & 4.23
& 2.86$\times$ & 4.35
& 2.75$\times$ & 4.11
& 2.62$\times$ & 4.00
& 2.36$\times$ & 3.56
& 1.87$\times$ & 2.72 \\
\quad +\textit{\SpdShort}
& \textbf{3.25}$\times$ & \textbf{5.10}
& \textbf{2.53}$\times$ & \textbf{3.83}
& \textbf{2.27}$\times$ & \textbf{3.39}
& \textbf{2.85}$\times$ & \textbf{4.80}
& \textbf{3.20}$\times$ & \textbf{5.02}
& \textbf{2.87}$\times$ & \textbf{4.45}
& \textbf{2.85}$\times$ & \textbf{4.46}
& \textbf{2.56}$\times$ & \textbf{3.97}
& \textbf{2.08}$\times$ & \textbf{3.04} \\
\midrule
Domino (A)
& 2.99$\times$ & 4.83
& 2.21$\times$ & 3.45
& 2.20$\times$ & 3.45
& 2.71$\times$ & 4.72
& 3.03$\times$ & 4.96
& 2.85$\times$ & 4.60
& 2.82$\times$ & 4.58
& 2.49$\times$ & 4.01
& 1.91$\times$ & 2.93 \\
\quad +\textit{\SpdShort}
& \textbf{3.35}$\times$ & \textbf{5.55}
& \textbf{2.52}$\times$ & \textbf{4.04}
& \textbf{2.32}$\times$ & \textbf{3.66}
& \textbf{2.89}$\times$ & \textbf{5.14}
& \textbf{3.24}$\times$ & \textbf{5.42}
& \textbf{2.97}$\times$ & \textbf{4.85}
& \textbf{3.01}$\times$ & \textbf{4.96}
& \textbf{2.67}$\times$ & \textbf{4.35}
& \textbf{2.08}$\times$ & \textbf{3.23} \\[-0.3ex]
\bottomrule
\end{tabular*}
\vspace{-6mm}
\end{table*}

%% file: sec/5_evaluation.tex
\section{Experiments}
\label{sec:experiments}

\subsection{Experimental Setup}
\label{subsec:setup}

\paragraph{Models}
We evaluate across three vision-language models: two GUI agent models, \texttt{GUI-Owl-1.5-8B-Think}~\cite{mobileagentv35} and \texttt{MAI-UI-8B}~\cite{maiui}, and one general-purpose VLM, \texttt{Qwen3-VL-8B-Thinking}~\cite{qwen3vl}. For each model, we train two diffusion drafters, DFlash~\cite{dflash} and Domino~\cite{domino}, in two stages. First, we train general-purpose (G) variants on 1.42M Open-PerfectBlend~\cite{openperfectblend} samples for 3 epochs. We then fine-tune them on 16K AndroidControl~\cite{androidcontrol} GUI samples for 2 epochs to obtain GUI-adapted (A) variants. These combinations span diverse deployment scenarios and drafter training domains, enabling a comprehensive evaluation of \SpdShort{}. Additional training details are provided in~\cref{app:training}.

\paragraph{Benchmarks}
We evaluate on three GUI navigation datasets: GUI-Odyssey~\cite{guiodyssey} and AITW~\cite{aitw} for mobile, and Mind2Web~\cite{mind2web} for web navigation. Evaluation is conducted offline with ground-truth screenshots and action histories.

\paragraph{Baselines}
For speculative decoding, we compare~\SpdShort{} against vanilla autoregressive decoding and two block-diffusion speculative decoding baselines, DFlash~\cite{dflash} and Domino~\cite{domino}. For visual token pruning, we compare against three pre-LLM pruning methods: DivPrune~\cite{divprune}, CDPruner~\cite{cdpruner}, and ZOO-Prune~\cite{zooprune}. For a controlled comparison, all visual token pruning methods are combined with \SpdShort{}. We retain 20\% or 10\% of historical visual tokens, while leaving the current screenshot unpruned.

\paragraph{Metrics}
For speculative decoding, we evaluate all methods using the mean acceptance length $\tau$ and decoding speedup over the autoregressive baseline. For visual token pruning, we report step accuracy following the evaluation protocol of SimpAgent~\cite{simpagent} at matched historical pruning ratio.

\paragraph{Experimental Settings}
All methods are evaluated using \texttt{SGLang 0.5.17}, with RadixAttention enabled for prefix caching, on a single NVIDIA A100 80GB GPU in BF16 precision. We use greedy decoding ($T=0$) for all methods and set the drafter block size to $B=16$. For \VtpLong{}, we use the GUI-adapted (A) drafters, and set the temporal weighting parameter $\gamma=0.2$. Our main results are measured at a request concurrency of 1 and a maximum generation length of 2,048 tokens. We provide implementation details in~\cref{app:implementation}.

\vspace{-1mm}

\input{tables/vtp_main}

\subsection{Main Results}
\label{subsec:mainresults}

\vspace{-2mm}

\paragraph{\SpdShort{} consistently improves acceptance length and translates to speedup}
As shown in~\cref{tab:spd},~\SpdShort{} substantially increases the acceptance length of state-of-the-art diffusion drafters. On average, it raises acceptance length from 2.45 to 3.50 for general-purpose drafters and from 3.95 to 4.40 for GUI-adapted ones. In particular,~\SpdShort{} improves acceptance length by up to 1.68 with Domino (G) for GUI-Owl on GUI-Odyssey. These gains translate directly into decoding speedup. On average,~\SpdShort{} accelerates decoding by 35.0\% over general-purpose drafters and 9.2\% over GUI-adapted ones. Notably, it reaches up to $3.35\times$ speedup over autoregressive decoding.

\paragraph{GUI fine-tuning shifts the gain from actions to reasoning}
General-purpose drafters underperform on GUI tasks due to training distribution mismatch, leaving them at an average acceptance length of 2.45. While fine-tuning on AndroidControl raises this to 3.95,~\SpdShort{} alone recovers 68\% of the tuning-induced reasoning gain and 61\% of the action gain without training. For GUI-adapted drafters, the action gain shrinks the most because predefined action formats can be easily memorised during fine-tuning. Regardless,~\RegexShort{} still improves the average action acceptance length from 5.97 to 6.16 with negligible overhead.~\SpdShort{}'s primary contribution therefore shifts to reasoning (\cref{tab:phase_breakdown}), increasing its share to saved verification steps from 38--76\% to 82--100\%.

\paragraph{\SpdShort{} retains action gains under domain mismatch}
While action gains for GUI-adapted drafters diminish in-domain, they persist under domain shifts. Fine-tuning yields the smallest average acceptance length gain on Mind2Web at 0.94, versus 2.24 on GUI-Odyssey and 1.30 on AITW. This stems from an action-space mismatch. Drafters tuned on AndroidControl learn touch actions like \texttt{click} and \texttt{swipe}, whereas Mind2Web requires unseen mouse actions like \texttt{left\_click} and \texttt{mouse\_move}. On Mind2Web,~\SpdShort{} still raises action acceptance length by up to 0.72 for GUI-Owl and Qwen3-VL, against at most 0.21 on mobile benchmarks. Real-world deployments vary widely in user requests and history compaction, making per-setup fine-tuning infeasible. \SpdShort{} bridges this gap without retraining.

\input{tables/spd_breakdown_gui_owl}

\paragraph{Drafter-guided historical pruning preserves accuracy when historical context is compressed}
As shown in~\cref{tab:vtp}, drafter-guided pruning loses only 1.2 and 1.8 percentage points (pp) on average at the 20\% and 10\% history retention ratios, compared with 2.3--3.2 and 3.1--3.9 pp for the baselines. Even with 90\% of the history pruned, our method with~\VtpLong{} preserves 96.8\% of the original accuracy on average. Notably, at a 20\% history retention ratio, Qwen3-VL with the Domino drafter matches its unpruned accuracy on AITW. Across all evaluated settings, both the DFlash and Domino variants rank first or second regardless of the drafter architecture. These results demonstrate that reusing existing drafter attention enables effective pre-LLM token removal without extra overhead, retaining the task awareness of in-LLM pruning.

\subsection{Analysis and Ablation Studies}
\label{subsec:analysis}

\paragraph{Ablation on Retrieval Sources}
As shown in~\cref{tab:spd_sources}, both intra- and inter-episode retrieval contribute to the gain of~\LookupShort{}. For DFlash (A), retrieving from the current episode raises reasoning acceptance length by 0.09--0.58 across benchmarks. Incorporating past episodes yields a further 0.14--0.39.~\cref{fig:spd_sources} shows how the total gain evolves as history accumulates. Within an episode, the gain grows gradually with the number of turns as earlier turns become retrievable, from 0.67 at the first turn to 1.07 at the tenth on GUI-Odyssey. Across episodes, it rises from 0.62 in the first seven to about 1.1 after 200 episodes, then saturates.

\input{tables/ablation_spd_sources}

\paragraph{Ablation on Block Allocation}
\cref{tab:spd_block_allocation} compares different block allocation strategies. Retrieval-only and Drafter-only fill the block solely with retrieved or draft tokens. Retrieval-first replaces the draft branch with the retrieved continuation upon a suffix match of at least two tokens. On the other hand, fixed allocation statically allocates $b_R$ block positions to the retrieved branch, using the best $b_R \in \{1, \dots, 15\}$ per benchmark. The oracle selects the optimal $b_R$ at each verification step given the target model's actual output. Both are infeasible to achieve in real deployment. Our survival-guided allocation consistently outperforms all non-oracle baselines. Notably, it achieves 96.6--99.3\% of the oracle.

\vspace{-2mm}

\input{tables/ablation_spd_block_allocation}

\paragraph{Ablation on Temporal Weighting}
\cref{tab:vtp_allocation} compares uniform ($\gamma=1$) and decaying ($\gamma=0.2$) temporal weights across historical screenshots under the same history retention ratio. Decaying weights improve accuracy by 0.3--0.6 pp across all six benchmarks. While drafter attention identifies important tokens within each screenshot, temporal decay provides complementary gains by allocating more budget to recent history. This prioritises recent screenshots that are most relevant to the current interaction turn.

\input{tables/ablation_vtp_breakdown}

\paragraph{Joint Prefill-Decoding Speedup}
\ours{} achieves a higher end-to-end speedup than accelerating either stage alone (\cref{fig:intro}c). Pruning alone yields only $1.16\times$ as decoding remains the bottleneck, whereas \SpdShort{} reaches $1.87\times$, thereby shifting the bottleneck to prefill. Through joint optimisation, \ours{} achieves a $2.37\times$ speedup by mitigating bottlenecks across both stages, outperforming DivPrune and DFlash alone by 104\% and 34.7\%, respectively.

\paragraph{Resource Overhead}
\cref{tab:overhead} shows that~\SpdLong{} adds 7.55\,ms per turn (0.36\% of model inference time). Updating the tool-call grammar automaton during verification is the largest component. In terms of memory, no additional GPU allocation is required. On the CPU, the compiled grammar and the current-episode retrieval state cost 56.4 and 0.6\,MB on average, while the past-episode corpus grows by 27.9\,MB per 100 episodes. Since retrieval gains saturate after a few hundred episodes, the corpus can be bounded by retaining only recent episodes.

\VtpLong{} adds 2.54\,ms per turn (0.34\%) for loading attention maps, allocating budgets, and selecting visual tokens. The additional forward pass to obtain the attention map adds no measurable latency at concurrency 1. This overhead slightly exceeds DivPrune and CDPruner but is substantially lower than ZOO-Prune (9.42\,ms). The memory footprint is also small: each attention map stores one FP16 value per visual token, and readout reuses the drafter's existing KV cache. Whereas other methods obtain model-aware pruning signals from early target-model layers or a separate small VLM~\cite{zhao2025sgl}, our method efficiently obtains task-relevant pruning signals from drafter attention. 

\vspace{-2mm}
\input{tables/ablation_overhead}

%% file: tables/vtp_main.tex
\begin{table*}[t]
\centering
\small
\setlength{\tabcolsep}{2pt}
\renewcommand{\arraystretch}{0.95}
\setlength{\aboverulesep}{0.5ex}
\setlength{\belowrulesep}{0.5ex}
\caption{
\textbf{Step accuracy with visual token pruning.}
We report step accuracy (\%) across three target models, retaining 20\% or 10\% of historical visual tokens and all current visual tokens. The best result at each budget is in \textbf{bold}, and the second best is \underline{underlined}.
}
\label{tab:vtp}
\vspace{-2mm}
\begin{tabularx}{\textwidth}{
  @{}
  l
  @{\hspace{0.4em}}
  l
  @{\hspace{0.1em}}
  *{9}{>{\centering\arraybackslash}X}
  @{}
}
\toprule
& & \multicolumn{3}{c}{
    \makebox[0pt][c]{\textbf{GUI-Owl-1.5-8B}}
  }
& \multicolumn{3}{c}{
    \makebox[0pt][c]{\textbf{MAI-UI-8B}}
  }
& \multicolumn{3}{c}{
    \makebox[0pt][c]{\textbf{Qwen3-VL-8B}}
  } \\
\cmidrule(l{0pt}r{3pt}){3-5}
\cmidrule(l{3pt}r{3pt}){6-8}
\cmidrule(l{3pt}r{0pt}){9-11}
\textbf{Retention} & \textbf{Method}
  & \makebox[0pt][c]{GUI-Ody}
  & \makebox[0pt][c]{AITW}
  & \makebox[0pt][c]{M2Web}
  & \makebox[0pt][c]{GUI-Ody}
  & \makebox[0pt][c]{AITW}
  & \makebox[0pt][c]{M2Web}
  & \makebox[0pt][c]{GUI-Ody}
  & \makebox[0pt][c]{AITW}
  & \makebox[0pt][c]{M2Web} \\[-0.3ex]
\midrule
\multicolumn{2}{@{}l}{Unpruned}
& 57.7 & 60.5 & 48.4
& 62.2 & 62.4 & 50.4
& 54.3 & 66.2 & 45.6 \\[-0.3ex]
\midrule
\multirow{5}{*}{\shortstack[l]{\textit{History 20\%}\\\textit{Current 100\%}}}
& DivPrune
& 54.0 & 56.5 & 45.6
& 60.9 & 60.5 & 48.9
& 52.5 & 64.9 & 43.4 \\
& CDPruner
& 52.1 & 55.9 & 44.0
& 59.9 & 60.2 & 48.5
& 51.2 & 64.5 & 42.5 \\
& ZOO-Prune
& 53.3 & 57.2 & 45.6
& 59.7 & 60.7 & 48.9
& 51.8 & 64.4 & 44.0 \\[-0.3ex]
\cmidrule{2-11}
& \textit{Drafter-Guided} (DFlash)
& \underline{55.3} & \textbf{58.1} & \underline{47.0}
& \textbf{61.8} & \textbf{61.6} & \textbf{49.7}
& \textbf{52.7} & \underline{66.1} & \underline{44.1} \\
& \textit{Drafter-Guided} (Domino)
& \textbf{55.4} & \textbf{58.1} & \textbf{47.1}
& \underline{61.6} & \textbf{61.6} & \textbf{49.7}
& \underline{52.6} & \textbf{66.2} & \textbf{44.3} \\[-0.3ex]
\midrule
\multirow{5}{*}{\shortstack[l]{\textit{History 10\%}\\\textit{Current 100\%}}}
& DivPrune
& 52.9 & 55.1 & 45.1
& 59.8 & 60.3 & 48.1
& 51.5 & 63.4 & 43.5 \\
& CDPruner
& 51.0 & 55.2 & 44.0
& 59.3 & 59.3 & 47.8
& 49.9 & 63.4 & 42.7 \\
& ZOO-Prune
& 51.9 & 55.0 & 45.0
& 58.8 & 59.9 & 48.4
& 50.8 & 63.4 & 43.6 \\[-0.3ex]
\cmidrule{2-11}
& \textit{Drafter-Guided} (DFlash)
& \textbf{54.5} & \underline{57.3} & \textbf{46.7}
& \textbf{61.0} & \textbf{61.4} & \textbf{49.1}
& \underline{51.7} & \textbf{65.9} & \underline{43.9} \\
& \textit{Drafter-Guided} (Domino)
& \underline{54.3} & \textbf{58.5} & \underline{46.5}
& \underline{60.9} & \underline{60.9} & \underline{49.0}
& \textbf{51.9} & \underline{65.5} & \textbf{44.2} \\[-0.3ex]
\bottomrule
\end{tabularx}
\vspace{-5mm}
\end{table*}

%% file: tables/spd_breakdown_gui_owl.tex
\newcommand{\deltaval}[2]{\makebox[0pt][r]{$#1$}#2}

\begin{table}[t]
\centering
\small
\setlength{\tabcolsep}{2pt}
\renewcommand{\arraystretch}{0.95}
\setlength{\aboverulesep}{0.5ex}
\setlength{\belowrulesep}{0.5ex}
\caption{
\textbf{Phase-wise breakdown of \SpdShort{}.}
Acceptance length improvements ($\Delta\tau$) and saved verification steps per turn of \SpdShort{} on GUI-Owl across reasoning ($\tau_r$) and action ($\tau_a$) phases.
}
\label{tab:phase_breakdown}
\vspace{-2mm}
\begin{tabularx}{\columnwidth}{
    @{}
    l
    @{\hspace{0.8em}}
    *{6}{>{\centering\arraybackslash}X}
    @{}
}
\toprule
& \multicolumn{2}{c}{\makebox[0pt][c]{\textbf{GUI-Ody}}}
& \multicolumn{2}{c}{\makebox[0pt][c]{\textbf{AITW}}}
& \multicolumn{2}{c}{\makebox[0pt][c]{\textbf{M2Web}}} \\
\cmidrule(l{0pt}r{3pt}){2-3}
\cmidrule(l{3pt}r{3pt}){4-5}
\cmidrule(l{3pt}r{0pt}){6-7}
Method
& $\tau_r$ & $\tau_a$
& $\tau_r$ & $\tau_a$
& $\tau_r$ & $\tau_a$ \\[-0.3ex]
\midrule
DFlash (G)
& 2.37 & 2.53 & 2.26 & 3.10 & 2.43 & 3.08 \\
\quad +\textit{\SpdShort}
& \textbf{3.90} & \textbf{4.94}
& \textbf{3.04} & \textbf{4.69}
& \textbf{2.74} & \textbf{4.24} \\
$\Delta\tau$ (\%)
& \deltaval{+}{64.6} & \deltaval{+}{95.3}
& \deltaval{+}{34.5} & \deltaval{+}{51.3}
& \deltaval{+}{12.8} & \deltaval{+}{37.7} \\
Saved steps
& 20.0 & 9.1 & 6.8 & 5.2 & 5.2 & 5.0 \\[-0.3ex]
\midrule
DFlash (A)
& 3.85 & 6.63 & 2.49 & 5.66 & 2.74 & 4.83 \\
\quad +\textit{\SpdShort}
& \textbf{4.82} & \textbf{6.79}
& \textbf{3.25} & \textbf{5.87}
& \textbf{2.97} & \textbf{5.06} \\
$\Delta\tau$ (\%)
& \deltaval{+}{25.2} & \deltaval{+}{2.41}
& \deltaval{+}{30.5} & \deltaval{+}{3.71}
& \deltaval{+}{8.39} & \deltaval{+}{4.76} \\
Saved steps
& 6.3 & 0.2 & 5.6 & 0.3 & 3.2 & 0.5 \\[-0.3ex]
\midrule
Avg. tokens
& 121 & 47 & 60 & 48 & 112 & 56 \\[-0.3ex]
\bottomrule
\end{tabularx}
\vspace{-5mm}
\end{table}

%% file: tables/ablation_spd_sources.tex
\begin{figure}[h]
\centering

\begin{minipage}{\linewidth}
    \centering
    \small
    \setlength{\tabcolsep}{2pt}
    \renewcommand{\arraystretch}{0.95}
    \setlength{\aboverulesep}{0.5ex}
    \setlength{\belowrulesep}{0.5ex}
    \captionof{table}{
    \textbf{Ablation of retrieval sources.}
    Reasoning-phase acceptance length ($\tau_r$) of GUI-Owl with DFlash (G) and (A) as turns in the current episode and past episodes are cumulatively added.
    }
    \label{tab:spd_sources}
    \vspace{-2mm}
    \begin{tabularx}{\linewidth}{
        @{}
        l
        @{\hspace{0.5em}}
        *{6}{>{\centering\arraybackslash}X}
        @{}
    }
        \toprule
        & \multicolumn{2}{c}{
            \makebox[0pt][c]{\textbf{GUI-Ody}}
            }
        & \multicolumn{2}{c}{
            \makebox[0pt][c]{\textbf{AITW}}
            }
        & \multicolumn{2}{c}{
            \makebox[0pt][c]{\textbf{M2Web}}
            } \\
        \cmidrule(l{0pt}r{3pt}){2-3}
        \cmidrule(l{3pt}r{3pt}){4-5}
        \cmidrule(l{3pt}r{0pt}){6-7}
        \textbf{Source}
            & (G) & (A) & (G) & (A) & (G) & (A) \\[-0.3ex]
        \midrule
        Baseline
            & 2.37 & 3.85 & 2.26 & 2.49 & 2.43 & 2.74 \\
        \hspace{0.4em}+ Current episode
            & 3.28 & 4.43 & 2.63 & 2.87 & 2.56 & 2.83 \\
        \hspace{0.4em}+ Past episodes
            & 3.90 & 4.82 & 3.04 & 3.25 & 2.74 & 2.97 \\[-0.3ex]
        \bottomrule
    \end{tabularx}
\end{minipage}
\par\nointerlineskip\vspace{2pt}

\includegraphics[width=\linewidth]{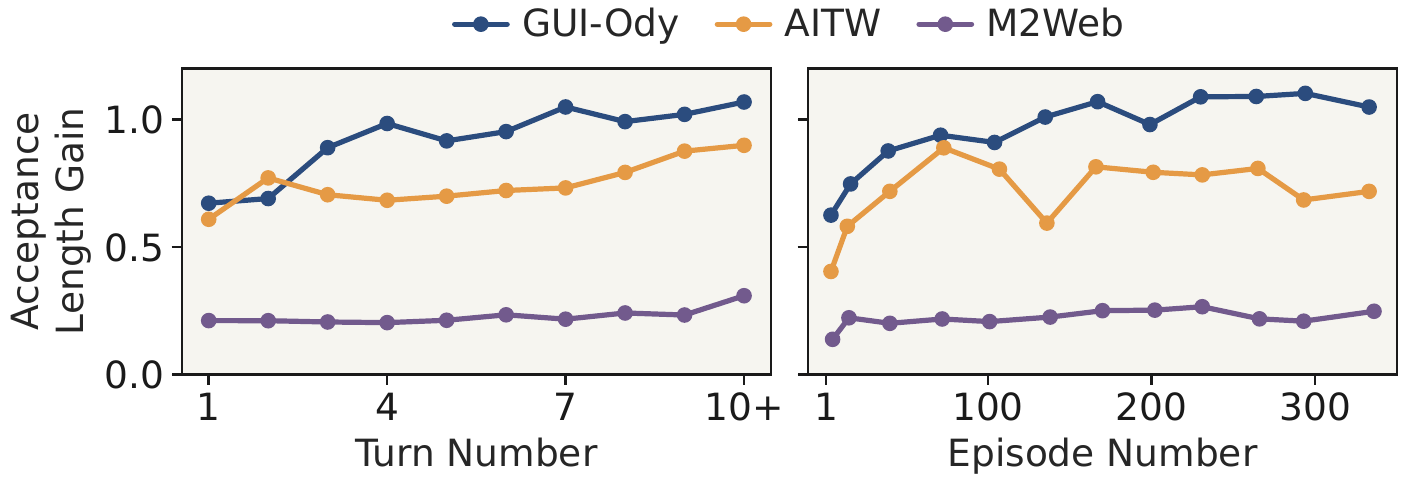}
\vspace{-7mm}
\caption{
$\tau_r$ gain of \SpdShort{} over DFlash (A) as the number of current-episode turns (left) and past episodes (right) grows.
}
\vspace{-3mm}
\label{fig:spd_sources}
\end{figure}

%% file: tables/ablation_spd_block_allocation.tex
\begin{table}[h]
\centering
\small
\setlength{\tabcolsep}{2pt}
\renewcommand{\arraystretch}{0.95}
\setlength{\aboverulesep}{0.5ex}
\setlength{\belowrulesep}{0.5ex}
\caption{
\textbf{Ablation of block allocation strategies.}
Reasoning-phase acceptance length ($\tau_r$) of GUI-Owl with DFlash (A) using $B=16$ (\% of Oracle). Oracle selects the optimal split per iteration.
}
\label{tab:spd_block_allocation}
\vspace{-2mm}
\begin{tabularx}{\columnwidth}{
    @{}
    l
    @{\hspace{0.1em}}
    *{3}{>{\centering\arraybackslash}X}
    @{}
}
    \toprule
    \textbf{Method}
    & \makebox[0pt][c]{\textbf{GUI-Ody}}
    & \makebox[0pt][c]{\textbf{AITW}}
    & \makebox[0pt][c]{\textbf{M2Web}} \\[-0.3ex]
    \midrule
    Retrieval only
        & 2.92\,(58.5\%)
        & 2.13\,(64.7\%)
        & 1.64\,(54.8\%) \\
    Drafter only
        & 3.85\,(77.2\%)
        & 2.49\,(75.7\%)
        & 2.74\,(91.6\%) \\
    Retrieval first
        & 3.32\,(66.5\%)
        & 2.44\,(74.2\%)
        & 1.99\,(66.6\%) \\
    Fixed allocation
        & 4.70\,(94.2\%)
        & 3.19\,(97.0\%)
        & 2.95\,(98.7\%) \\
    \textit{Survival-Guided}
        & \textbf{4.82}\,(96.6\%)
        & \textbf{3.25}\,(98.8\%)
        & \textbf{2.97}\,(99.3\%) \\[-0.3ex]
    \midrule
    Oracle
        & 4.99\,(100\%)
        & 3.29\,(100\%)
        & 2.99\,(100\%) \\[-0.3ex]
    \bottomrule
\end{tabularx}
\vspace{-4mm}
\end{table}

%% file: tables/ablation_vtp_breakdown.tex
\begin{table}[h]
\centering
\small
\setlength{\tabcolsep}{1pt}
\renewcommand{\arraystretch}{0.95}
\setlength{\aboverulesep}{0.5ex}
\setlength{\belowrulesep}{0.5ex}
\caption{
\textbf{Ablation of temporal weighting.}
Step accuracy (\%) of drafter-guided pruning (DFlash) averaged over three target models. We compare uniform ($\gamma = 1$) and decaying ($\gamma = 0.2$) weights at history retention ratios of 20\% and 10\%.
}
\label{tab:vtp_allocation}
\vspace{-2mm}
\begin{tabularx}{\columnwidth}{
  @{}
  l
  @{\hspace{0.1em}}
  *{6}{>{\centering\arraybackslash}X}
  @{}
}
  \toprule
  & \multicolumn{2}{c}{
      \makebox[0pt][c]{\textbf{GUI-Ody}}
    }
  & \multicolumn{2}{c}{
      \makebox[0pt][c]{\textbf{AITW}}
    }
  & \multicolumn{2}{c}{
      \makebox[0pt][c]{\textbf{M2Web}}
    } \\
  \cmidrule(l{0pt}r{3pt}){2-3}
  \cmidrule(l{3pt}r{3pt}){4-5}
  \cmidrule(l{3pt}r{0pt}){6-7}
  \textbf{History} & 20\% & 10\% & 20\% & 10\% & 20\% & 10\% \\[-0.3ex]
  \midrule
  Uniform ($\gamma=1$)
      & 56.0 & 55.1 & 61.5 & 61.2 & 46.4 & 46.1 \\
  Decaying ($\gamma=0.2$)
      & \textbf{56.6} & \textbf{55.7}
      & \textbf{61.9} & \textbf{61.5}
      & \textbf{46.9} & \textbf{46.6} \\[-0.3ex]
  \bottomrule
\end{tabularx}
\vspace{-4mm}
\end{table}

%% file: tables/ablation_overhead.tex
\begin{table}[h]
\centering
\small
\setlength{\tabcolsep}{1pt}
\renewcommand{\arraystretch}{0.95}
\setlength{\aboverulesep}{0.5ex}
\setlength{\belowrulesep}{0.5ex}
\caption{
\textbf{Latency and memory overhead of (a) \SpdShort{} and (b) VTP methods.} Latency is in ms/turn (\% of model time), and memory is in MB; memory for past episodes is per 100 episodes.
}
\label{tab:overhead}
\vspace{-2mm}
\renewcommand{\thesubtable}{(\alph{subtable})}

\begin{subtable}[t]{\dimexpr0.52\linewidth-0.33em\relax}
    \centering
    \vspace{0pt}
    \phantomsubcaption
    \label{tab:spd_overhead}
    \begin{tabularx}{\linewidth}{
        @{}
        l
        @{\hspace{0.6em}}
        l
        @{\hspace{0.3em}}
        >{\raggedleft\arraybackslash}X
        @{}
    }
        \toprule
        \multicolumn{3}{@{}l@{}}{
            \textbf{\thesubtable}\ \SpdShort{}
        } \\[-0.3ex]
        \midrule
        \multicolumn{2}{@{}l}{\textbf{Metric}}
            & \textbf{Overhead} \\[-0.3ex]
        \midrule
        \multicolumn{2}{@{}l}{Latency}
            & \makebox[0pt][r]{7.55\,(0.36\%)} \\[-0.3ex]
        \midrule
        \multicolumn{2}{@{}l}{GPU}
            & 0\,MB \\[-0.3ex]
        \midrule
        \multirow{3}{*}{CPU}
            & \multirow{3}{*}{%
                \makebox[0pt][r]{%
                    \rule{0.4pt}{2.1\baselineskip}%
                    \hspace{0.4em}%
                }%
              }Current episode
            & 0.6\,MB \\[-0.3ex]
            & Past episodes
            & 27.9\,MB \\[-0.3ex]
            & Grammar
            & 56.4\,MB \\[-0.3ex]
        \bottomrule
    \end{tabularx}
\end{subtable}\hfill
\begin{subtable}[t]{\dimexpr0.48\linewidth-0.27em\relax}
    \centering
    \vspace{0pt}
    \phantomsubcaption
    \label{tab:vtp_overhead}
    \def\vtprowstrut{%
        \vrule width 0pt
        height 1.36\ht\strutbox
        depth 1.36\dp\strutbox\relax
    }
    \begin{tabularx}{\linewidth}{
        @{}
        >{\raggedright\arraybackslash}X
        @{\hspace{0.1em}}
        c
        @{}
    }
        \toprule
        \multicolumn{2}{@{}l@{}}{
            \textbf{\thesubtable}\ VTP methods
        } \\[-0.3ex]
        \midrule
        \textbf{Method} & \textbf{Latency} \\[-0.3ex]
        \midrule
        \vtprowstrut DivPrune
            & 1.11(0.15\%) \\[-0.3ex]
        \vtprowstrut CDPruner
            & 1.84(0.25\%) \\[-0.3ex]
        \vtprowstrut ZOO-Prune
            & 9.42(1.25\%) \\[-0.3ex]
        \vtprowstrut \mbox{\textit{Drafter-Guided}}
            & 2.54(0.34\%) \\[-0.3ex]
        \bottomrule
    \end{tabularx}
\end{subtable}
\end{table}

\vspace{-5mm}

%% file: sec/6_conclusion.tex
\section{Conclusion}
\label{sec:conclusion}

We present~\ours{}, a training-free, plug-and-play framework for efficient VLM-based GUI agent inference. On the decoding side, we exploit the asymmetry between reasoning and action phases with~\SpdLong{}, achieving up to $3.35\times$ decoding speedup over autoregressive baselines. For prefill acceleration, we observe that acceptance length is robust to pruning historical screenshots but sensitive to pruning the current one. We further propose~\VtpLong{} to guide historical pruning with drafter attention, retaining 96.8\% of baseline accuracy on average with 90\% of historical tokens pruned. Beyond GUI agents, phase-aware speculative decoding can broadly benefit agentic workloads that alternate between free-form reasoning and structured actions. Our work demonstrates that co-designing prefill and decoding optimisations enables full-pipeline efficient agent inference.

\clearpage

%% file: sec/X_suppl.tex
\clearpage
\setcounter{page}{1}
\maketitlesupplementary
\appendix
\crefalias{section}{appendix}
\crefalias{subsection}{subappendix}
\crefalias{subsubsection}{subappendix}

\section{Extended Related Work}
\label{app:related}

\subsection{GUI Agents}

Graphical User Interface (GUI) agents autonomously map screen observations and user instructions to executable UI actions across mobile, web, and desktop environments~\cite{mind2web, aitw, guiodyssey}. While early agents relied on platform-specific metadata such as HTML or accessibility trees, recent advances in vision-language models (VLMs)~\cite{flamingo, blip2, llava} have enabled vision-based agents that operate directly on raw screenshots~\cite{seeclick, cogagent}. Building on this paradigm, modern GUI agents unify perception, reasoning, and action within a single model trained on large-scale trajectory corpora and reinforcement learning~\cite{showui, uitars, uitars2, uivenus2, mobileagentv35, maiui}.

GUI agents operate in multi-turn interaction loops, where each turn conditions on a high-resolution current screenshot for fine-grained coordinate grounding alongside accumulated historical screenshots and actions. Consequently, visual tokens dominate the input, imposing substantial prefill overhead. Meanwhile, reasoning-enabled agents generate free-form chain-of-thought (CoT) traces prior to emitting schema-conforming action commands~\cite{uitars,mobileagentv35,maiui}. This reasoning phase improves planning on complex interactive tasks, but incurs significant autoregressive decoding overhead. To address both latency bottlenecks, this work co-designs speculative decoding and visual token pruning to jointly accelerate the prefill and decoding stages.

\vspace{-2mm}

\subsection{Speculative Decoding}

Speculative decoding reduces decoding latency without altering the output distribution~\cite{leviathan2023fast,chen2023accelerating}. Specifically, a lightweight drafter proposes candidate tokens that the target model verifies in a single forward pass. Drafters span several architectures, including standalone smaller models~\cite{leviathan2023fast,chen2023accelerating}, autoregressive feature-level drafters~\cite{li2024eagle,li2024eagle2,eagle3}, parallel multi-head predictors~\cite{medusa}, and diffusion-based models~\cite{christopher2025speculative}. 

Beyond drafter architecture, orthogonal optimisations target decoding efficiency and long-context bottlenecks. SpecVocab~\cite{specvocab} restricts drafting to a compact speculative vocabulary, Sequoia~\cite{sequoia} builds hardware-aware draft trees, and AdaptiveSpec~\cite{marginsnotwindows} relaxes per-step acceptance criteria for lossy acceleration. For long-context workloads, TriForce~\cite{triforce} introduces hierarchical speculative decoding.

Recently, block-diffusion speculative decoding has emerged, predicting a block of $B$ draft tokens in a single drafter forward pass. DFlash~\cite{dflash,dflash2} drafts blocks in parallel with a diffusion model, and Domino~\cite{domino} adds a low-rank GRU head to model intra-block dependencies. Extending this paradigm, DSpark~\cite{dspark} adapts the verification length based on prefix survival probabilities, DDTree~\cite{ddtree} constructs block-diffusion draft trees, and WhiFlash~\cite{whiflash} routes tokens between autoregressive and diffusion drafters.

Extending speculative decoding to multimodal LLMs introduces visual-alignment challenges and substantial visual-context overhead~\cite{mmspec,mmspec_diffusion_survey}. Training-based methods such as MASSV~\cite{massv} connect the target vision encoder to a smaller language model and use self-data distillation to align token distributions. ViSpec~\cite{vispec} compresses image tokens with a lightweight vision adaptor and injects them into draft hidden states. Training-free methods instead optimise test-time execution. For video LLMs, SpecVLM~\cite{ji25specvlm} prunes redundant video tokens for the drafter using target-model attention. 

Retrieval-based methods exploit recurring patterns in prompts and histories, making them particularly effective for agentic workloads~\cite{samdecoding,agentspec,copyspec,mobilespec_npu}. SuffixDecoding~\cite{suffixdecoding} builds suffix structures over ongoing and historical outputs to retrieve candidate continuations at negligible cost. Graft~\cite{graft} combines retrieved and neural draft tokens within a hybrid candidate tree. Targeting structured tool execution, ToolSpec~\cite{toolspec} pairs schema-aware constraints with retrieval over past invocations, and OoO-Spec~\cite{ooospec} applies out-of-order semantic speculation to tool calling.

Prior retrieval-based works apply a uniform strategy across the entire output. In this work, we identify the asymmetry between long, freeform reasoning and short, structured actions, and propose phase-specialised optimisations. Specifically, SuffixDecoding~\cite{suffixdecoding} switches exclusively between suffix retrieval and an autoregressive drafter using a score threshold. In contrast, \LookupLong{} (\LookupShort{}) branches both proposals at first divergence, allocating block positions by survival probability without thresholds. Other works like Graft~\cite{graft} replace low-confidence autoregressive tree branches with retrieval to improve acceptance length. When a single diffusion pass generates the full block, this saving vanishes. For block-diffusion drafters, Graft avoids tree verification by appending retrieval after a confident prefix. Consequently, any early prefix error invalidates the chained tail, leaving block inefficiency unresolved. Furthermore, Graft overlooks retrieval reliability and requires warm-up with per-dataset calibration. In comparison, \LookupShort{} verifies both branches in parallel within a single block under causal attention. This prevents early draft errors from invalidating the retrieved branch, while survival-guided block allocation requires neither warm-up nor tuning.

Another line of work, including ToolSpec~\cite{toolspec} and OoO-Spec~\cite{ooospec}, targets structured tool invocations. Specifically, ToolSpec relies on retrieval and drafter-free FSM schema filling, which only emits deterministic syntax and enumerates small candidate sets, failing on visual coordinates. In contrast, \RegexLong{} (\RegexShort{}) couples grammar constraints with a trained drafter rather than relying on static enumeration. \RegexShort{} revises the drafter's proposals by forcing deterministic tokens and selecting the highest-logit valid token for constrained fields, thereby preserving visual grounding for action types and coordinates.

\vspace{-2mm}

\subsection{Visual Token Pruning}

Visual token pruning reduces multimodal prefill latency by discarding redundant tokens~\cite{fastv,divprune}. Existing methods fall into in-LLM and pre-LLM paradigms. In-LLM methods drop tokens inside transformer layers using intermediate attention, preserving task-relevant regions. For example, FastV~\cite{fastv} removes low-attention tokens after early layers, and SparseVLM~\cite{sparsevlm} progressively sparsifies them with text-visual attention. However, this incurs full early-layer computation and breaks prefix caching in serving engines~\cite{vllm,sglang}. To preserve cache compatibility, pre-LLM methods prune tokens before LLM prefill. Within this paradigm, DivPrune~\cite{divprune} selects a diverse visual subset, CDPruner~\cite{cdpruner} maximises instruction-conditioned diversity, and ZOO-Prune~\cite{zooprune} scores tokens with zeroth-order gradients. Operating outside the LLM, these methods nonetheless lack the task awareness captured by LLM attention.

Recent works extend visual token pruning to multi-turn GUI agents, where high-resolution screenshots and growing histories incur substantial prefill cost~\cite{histprune}. GUIPruner~\cite{guipruner} applies temporally decaying downscaling to historical frames alongside current-frame pruning. AQuaUI~\cite{aquaui} allocates tokens using adaptive quadtrees, and FocusUI~\cite{ouyang2026focusui} performs position-preserving token selection for UI grounding. Beyond a single optimisation technique, we investigate the interaction between speculative decoding and visual token pruning for full-pipeline acceleration. 

Prior to our work, SGL~\cite{zhao2025sgl} employs an auxiliary small VLM to guide visual token pruning for a larger target model. Beyond substantial memory and critical-path latency overhead, it requires a same-family counterpart and risks attention mismatch when the small model diverges from the target. In contrast, \ours{} reuses earlier-turn attention scores from the deployed, target-aligned drafter to prune historical visual tokens prior to LLM prefill, preserving both task awareness and prefix caching with negligible overhead.

\section{Limitations and Future Work}
\label{app:limitations}

Our evaluation covers two GUI-specific models (GUI-Owl-1.5-8B-Think and MAI-UI-8B) and one general VLM (Qwen-3-VL-8B-Think), all at the 8B scale. Although no component of \ours{} is tied to model size, changing model scale shifts the balance between prefill and decoding latency. We expect \ours{} to remain effective across scales. Even when one phase dominates, speculative decoding accelerates generation and historical pruning reduces prefill cost. 

A second limitation concerns the scope of drafter-guided pruning. This mechanism relies on earlier-turn attention and cannot guide pre-LLM pruning of the current screenshot. Consequently, single-turn queries and the initial step of a trajectory do not benefit from visual pruning.

In this paper, we focus on co-designing speculative decoding with visual token pruning. A natural extension is to integrate \ours{} with other compression techniques, including vision caching~\cite{vlcache,vlacache}, KV cache compression~\cite{guikv,starkv}, and model quantisation. These methods can further reduce latency and memory footprint over long trajectories. 

\vspace{-4mm}

\section{Algorithms}
\label{app:algorithms}

\vspace{-1mm}

This section details the auxiliary functions invoked within \SpdLong{}~(\cref{alg:pssd}), together with the prefill-stage visual token pruning algorithm for \VtpLong{}~(\cref{alg:vtp}).

Given a multimodal prompt $\mathbf x$, \cref{alg:pssd} (\textsc{Decode}) performs target prefill followed by iterative \SpdLong{} decoding across the reasoning and action phases. During the reasoning phase, each verification iteration invokes three \LookupShort{} functions. Specifically, \cref{alg:retrieve} (\textsc{Retrieve}) performs hierarchical suffix matching and returns the most recent continuation $\mathbf r$ corresponding to the longest matched suffix of length $n$. Subsequently, \cref{alg:allocate} (\textsc{Allocate}) partitions the remaining block positions between the draft and retrieved branches using estimated branch survival probabilities to construct the candidate tree $\mathcal T$. Following tree verification, \cref{alg:update_acceptance} (\textsc{UpdateAcceptance}) compares the retrieved branch against the committed tokens $\mathbf a$ to update the acceptance and verification counts $(H_{n,i}, N_{n,i})$ for subsequent block allocations. During the action phase, \cref{alg:pssd} omits retrieval and applies \RegexShort{} directly, revising draft tokens against the schema automaton prior to verification.

\Cref{alg:vtp} (\textsc{PruneHistory}) prunes historical visual tokens prior to prefill to construct the multimodal prompt $\mathbf x$, with the current screenshot unpruned. It distributes the history retention ratio $b$ across historical screenshots with temporal weighting and retains the highest-scoring tokens according to recorded drafter attention scores.

\input{algorithms/PSSD_auxiliary}

\vspace{-5mm}

\section{Drafter Training Details}
\label{app:training}

We train DFlash~\cite{dflash} and Domino~\cite{domino}, two families of block-diffusion drafters, for three target models. These include two GUI agents, \texttt{GUI-Owl-\allowbreak 1.5-\allowbreak 8B-Think}~\cite{mobileagentv35} and \texttt{MAI-UI-8B}~\cite{maiui}, and one general-purpose VLM, \texttt{Qwen3-VL-\allowbreak 8B-Thinking}~\cite{qwen3vl}. We initially train general-purpose (G) drafter variants on $1.42\text{M}$ text-only samples from Open-PerfectBlend~\cite{openperfectblend}. Subsequently, we fine-tune them on $16\text{K}$ GUI samples from AndroidControl~\cite{androidcontrol} to obtain the corresponding GUI-adapted (A) variants.

\subsection{Training Setup}

\paragraph{Training Data and Response Regeneration}
Following DFlash~\cite{dflash}, we regenerate all supervised assistant responses using the corresponding target model, sampling with each model's recommended decoding configuration:

\begin{itemize}
    \item \textbf{GUI-Owl:} Temperature = $0.6$, top-$p$ = $0.95$, top-$k$ = $40$.
    \item \textbf{Qwen3-VL:} Temperature = $1.0$, top-$p$ = $0.95$, top-$k$ = $20$.
    \item \textbf{MAI-UI:} Greedy decoding.
\end{itemize}

Since Open-PerfectBlend contains only text, the (G) drafters never observe visual tokens during training. We therefore perform GUI-adapted (A) fine-tuning to align them with in-domain multimodal GUI data. To construct the (A) training set, we use $16\text{K}$ single-step GUI samples from $3{,}000$ episodes of the AndroidControl~\cite{androidcontrol} training split and format each with the corresponding target model's native GUI prompt template. Each sample includes the current screenshot and up to $4$ historical screenshots. Across all three models, historical turns retain only ground-truth actions without prior reasoning traces, whilst only the final assistant turn is regenerated and supervised during training.

\paragraph{Architecture and Training Recipe}
Following the default configurations of DFlash~\cite{dflash} and Domino~\cite{domino}, all drafters use a $5$-layer transformer backbone with block size $B=16$, extracting target hidden states from layers $[1, 9, 17, 25, 33]$. Additionally, Domino employs a projector with GRU hidden dimension $1{,}024$ and embedding dimension $256$, together with a base-loss weight $\lambda_{\text{base}}$ linearly annealed from $1$ to $0$.

Both variants adopt AdamW with zero weight decay, gradient clipping at $1.0$, a cosine learning-rate schedule with warm-up ratio $0.04$, $512$ anchor positions per sequence, loss decay factor $\gamma_{\text{loss}}=7.0$, and BF16 precision:

\begin{itemize}
    \item \textbf{General-purpose (G):} Learning rate $6\times 10^{-4}$, $3$ epochs.
    \item \textbf{GUI-adapted (A):} Weights-only warm start from the (G) checkpoint, learning rate $1.5\times 10^{-4}$, 2 epochs.
\end{itemize}

\paragraph{Implementation}

Our training pipeline builds on SpecForge~\cite{li2026specforge}, using \texttt{SGLang 0.5.14} for target hidden-state capture and \texttt{SGLang 0.5.17} for serving evaluation. To support multi-turn GUI trajectory training for both DFlash and Domino, we extend SpecForge with multi-image VLM sample collation and reasoning-content parsing.

\paragraph{Training Cost}
All training runs use a single node of $8\times$ NVIDIA A100-80GB GPUs. General-purpose (G) training requires ${\sim}470$--$1{,}120$ GPU-hours per drafter depending on the target model, whereas GUI-adapted (A) fine-tuning is lightweight, requiring only ${\sim}29$--$33$ GPU-hours per drafter.

\subsection{Analysis of Drafter Training}

\begin{figure}[h]
    \centering
    \includegraphics[width=\linewidth]{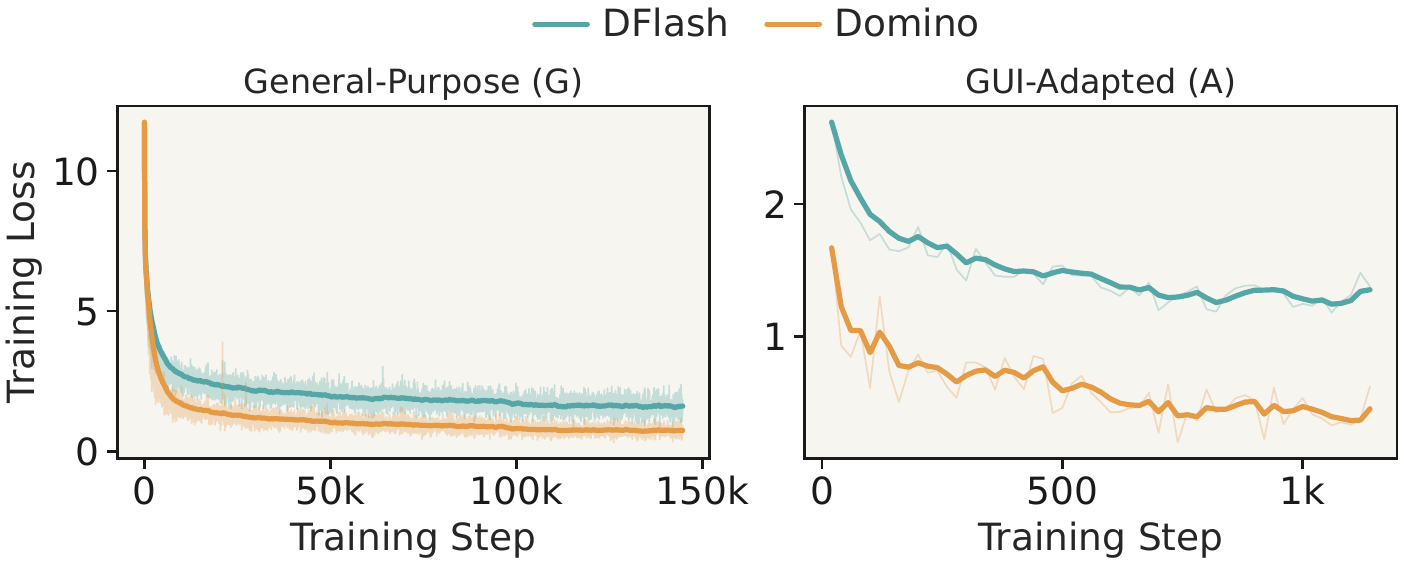}
    \caption{Training loss curves for GUI-Owl in general-purpose (G) training and GUI-adapted (A) fine-tuning.}
    \label{fig:drafter_loss_curves}
    \vspace{-5mm}
\end{figure}

\paragraph{Training Loss Dynamics}
\Cref{fig:drafter_loss_curves} shows the training loss trajectories for GUI-Owl across general-purpose (G) training and GUI-adapted (A) fine-tuning. During general-purpose training, both losses drop sharply in the initial steps and converge over 3 epochs. During GUI-adapted fine-tuning, the loss declines rapidly within the first 500 steps and largely stabilises by 1K steps (2 epochs). This indicates that a pre-trained block-diffusion drafter can quickly align with recurring in-domain GUI patterns through lightweight fine-tuning. However, rapid loss convergence alone does not address block inefficiency during inference or generalisation gaps on out-of-domain trajectories. Across the two families, absolute loss magnitudes are not directly comparable, as Domino's objective includes an auxiliary backbone loss whose weight $\lambda_{\text{base}}$ anneals linearly from 1 to 0 over each run.

\paragraph{In-Domain GUI Supervision}
Prior multimodal speculative decoding methods typically enhance visual drafting using generic vision-language instruction data, as in MSD~\cite{msd} and MASSV~\cite{massv}. For GUI agents, however, generic multimodal supervision is insufficient. As shown in \Cref{tab:spd_stage2_data}, fine-tuning the GUI-Owl DFlash (G) drafter on LLaVA-Instruct-150K~\cite{llava} slightly degrades performance on GUI-Odyssey, lowering the mean acceptance length $\tau$ from 2.41 to 2.37. In contrast, fine-tuning on just 16K in-domain AndroidControl~\cite{androidcontrol} samples raises $\tau$ by 1.84 and increases decoding speedup to $2.79\times$. Thus, effective GUI drafting depends on domain alignment rather than data volume. A compact set of in-domain GUI samples suffices to capture the target model's reasoning-and-action distribution, whereas much larger generic multimodal datasets offer little benefit.

\begin{table}[h]
\centering
\small
\setlength{\tabcolsep}{2pt}
\renewcommand{\arraystretch}{0.95}
\setlength{\aboverulesep}{0.5ex}
\setlength{\belowrulesep}{0.5ex}
\caption{
\textbf{Ablation of fine-tuning data.}
Mean acceptance length ($\tau$) and decoding speedup of GUI-Owl DFlash on GUI-Odyssey after fine-tuning (G) on LLaVA-Instruct-150K or AndroidControl.
}
\label{tab:spd_stage2_data}
\vspace{-2mm}
\begin{tabularx}{\linewidth}{
    @{}
    l
    @{\hspace{0.5em}}
    *{3}{>{\centering\arraybackslash}X}
    @{}
}
    \toprule
    \textbf{Drafter}
        & $\tau$ & $\Delta\tau$ & Speedup \\[-0.3ex]
    \midrule
    General-purpose (G)
        & 2.41 & -- & 1.66$\times$ \\
    \hspace{0.4em}+ General multimodal
        & 2.37 & $-0.04$ & 1.65$\times$ \\
    \hspace{0.4em}+ In-domain GUI (A)
        & 4.25 & $+1.84$ & 2.79$\times$ \\[-0.3ex]
    \bottomrule
\end{tabularx}
\vspace{-3mm}
\end{table}

\paragraph{Multi-Turn Supervision}
GUI agent trajectories are multi-turn, accumulating historical screenshots and past actions as a task progresses. We found that fine-tuning a drafter on single-turn inputs without history degrades the mean acceptance length $\tau$ during multi-turn inference as the context length grows. In line with the long-context mitigation discussed in DFlash~\cite{dflash}, our GUI-adapted (A) training set mixes samples across different steps within each episode, ranging from 0 (current screen only) to 4 historical screenshots. In this way, the drafter is exposed to varying history lengths, thereby supporting consistent drafting performance across both early steps and extended multi-turn horizons.

\section{Implementation and Evaluation Details}
\label{app:implementation}

\subsection{Common Evaluation Setup}

\paragraph{Inference Setup}
All evaluations run on \texttt{SGLang 0.5.17} using a single NVIDIA A100 80GB GPU in BF16 precision. Main results are measured at request concurrency 1, context length 32,768, and static memory fraction 0.70. We use greedy decoding ($T=0$, top-$k$=1) for both target models and drafters, with a maximum generation length of 2,048 tokens. RadixAttention is enabled for prefix caching. LLM prefill, drafter execution, and target verification use the FlashInfer backend (SGLang's A100 default), whereas the ViT uses \texttt{sgl-kernel}'s SM80 FlashAttention-3 kernel to avoid the non-Hopper Triton fallback. Episodes are evaluated offline with ground-truth history.

\paragraph{Benchmarks and Evaluation Protocols}
We evaluate on the GUI-Odyssey~\cite{guiodyssey} v2 \texttt{random\_split} test set, as well as the SeeClick~\cite{seeclick} splits of AITW~\cite{aitw} and pure-vision Mind2Web~\cite{mind2web} following SimpAgent~\cite{simpagent}. Each target model uses its official agent prompt and action space. GUI-Owl adopts the \texttt{X-PLUG/MobileAgent} v3.5 prompts (\texttt{mobile\_use} and \texttt{computer\_use}) for mobile and web tasks. MAI-UI adopts the \texttt{Tongyi-MAI/MAI-UI} prompt, with the AndroidWorld application list replaced by web action definitions for Mind2Web. Qwen3-VL uses the Qwen3-VL cookbook mobile prompt together with the OSWorld web prompt and multi-turn layout.

Following SimpAgent's visual history setting, we retain up to 4 historical screenshots on GUI-Odyssey and AITW, and 2 on Mind2Web. For earlier turns, only the textual action history is kept. We report step success rate (Step SR) under the SimpAgent evaluation protocol, where a step succeeds if the predicted action type, arguments, and grounding match the ground truth. Specifically, each step is evaluated using coordinate distance on GUI-Odyssey, screen-wise action matching on AITW, and bounding-box containment with matching operation and value on Mind2Web.

\subsection{\SpdLong{} Setup}

All \SpdShort{} results use the native resolution of the dataset screenshots, rounding them only to the 32~px token grid. We measure the mean acceptance length $\tau$ as the average number of accepted tokens per verification iteration across turns. For the reasoning-phase ($\tau_r$) and action-phase ($\tau_a$) acceptance length, each iteration is assigned to its starting phase. Decoding efficiency is reported via decoding speedup, computed as the ratio of pooled time per output token (TPOT) under autoregressive decoding to that under speculative decoding.

\subsection{\VtpLong{} Setup}

To ensure a fair comparison, all \VtpShort{} experiments follow the commonly adopted image preprocessing protocol of SimpAgent~\cite{simpagent,guipruner,histprune} for both historical and current screenshots. Specifically, the longest edge is resized to 512~px on GUI-Odyssey and AITW, whereas Mind2Web crops remain within the default bounds. We evaluate pruning methods by Step SR and history retention ratio. The latter denotes the fraction of retained post-merger historical visual tokens relative to unpruned input at the same resolution. We evaluate two retention ratios (20\% and 10\%) whilst keeping the current screenshot unpruned. Each \VtpShort{} run extracts importance scores from its own GUI-adapted (A) drafter.

All three baselines, including DivPrune~\cite{divprune}, CDPruner~\cite{cdpruner}, and ZOO-Prune~\cite{zooprune}, are evaluated under the same history retention ratios. For a controlled comparison, all visual token pruning methods are combined with \SpdShort{}. For the baselines, we follow their official settings. DivPrune selects tokens via pairwise cosine distances between visual tokens. CDPruner adopts its Qwen2.5-VL recipe, scoring tokens by cosine similarity to the mean goal embedding. ZOO-Prune uses 64 directions with a step size of 0.01.

\section{Additional Analysis}
\label{app:additional_analysis}

\subsection{Multi-Concurrency Results}

\begin{figure}[t]
    \centering
    \includegraphics[width=\linewidth]{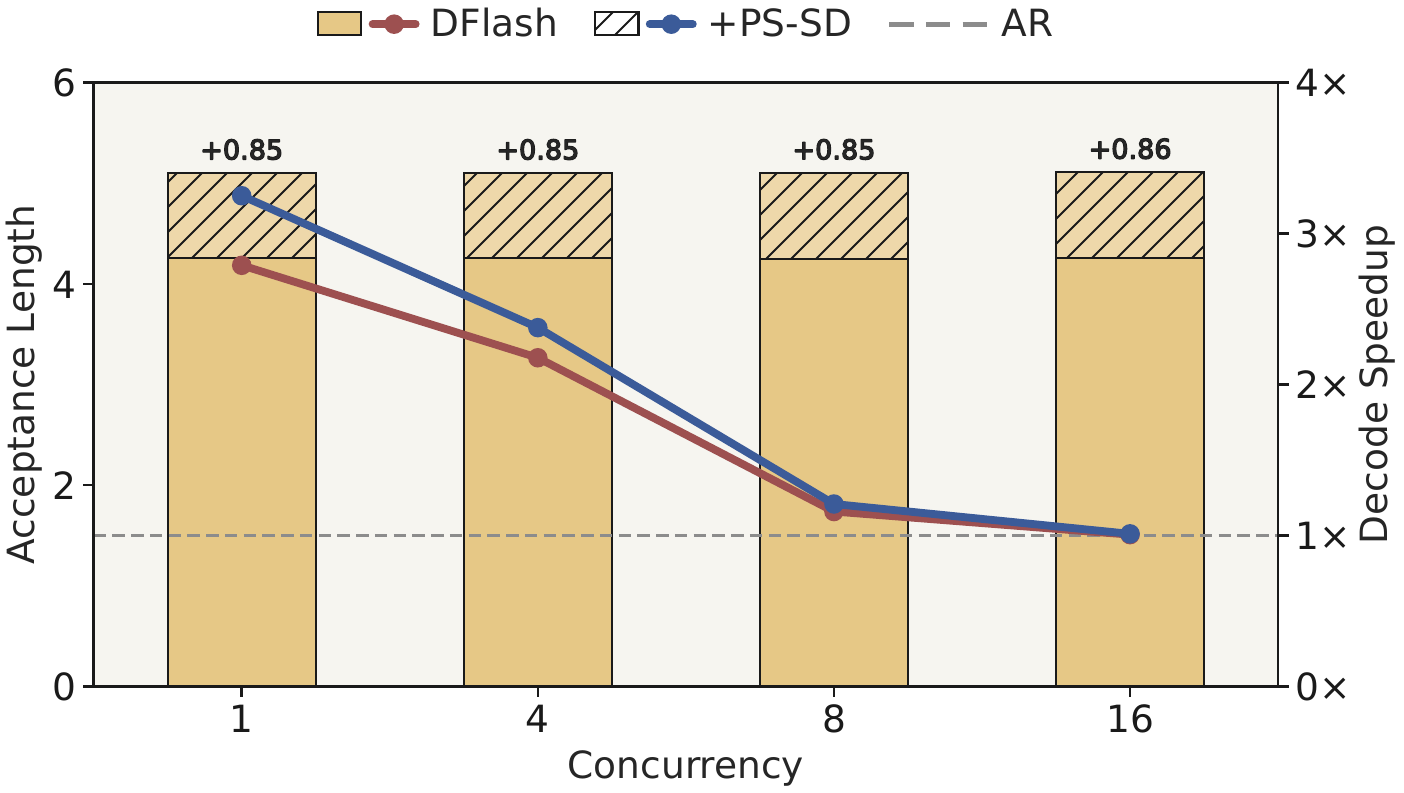}
    \caption{
    \textbf{Multi-concurrency scaling.}
    Mean acceptance length $\tau$ (bars; hatched regions denote the \SpdShort{} gain over DFlash) and decoding speedup over autoregressive decoding (lines) across concurrency levels 1, 4, 8, and 16.
    }
    \label{fig:concurrency}
    \vspace{-4mm}
\end{figure}

\Cref{fig:concurrency} evaluates decoding speedup and mean acceptance length $\tau$ at concurrency levels 1, 4, 8, and 16. As concurrency increases, the decoding speedup of both plain block-diffusion speculative decoding (DFlash) and \SpdShort{} decreases. This trend reflects an inherent batching trade-off of block-diffusion drafters as target verification becomes compute-bound, consistent with the multi-concurrency results reported in DFlash~\cite{dflash}. Crucially, \SpdShort{}'s acceptance length improvement remains invariant to concurrency, consistently delivering higher decoding speedup than DFlash. Beyond verification batching, multi-turn GUI workloads also incur heavy visual prefill across current and historical screenshots. With chunked and mixed prefill disabled, each prefill stalls all ongoing decodes and further slows decoding. This bottleneck highlights the importance of co-designing \SpdShort{} and \VtpShort{}. Pruning historical visual tokens reduces prefill overhead, directly complementing decoding acceleration.

\subsection{Phase-Wise Breakdown Across Target Models and Drafters}

\Cref{tab:phase_breakdown} presents the full phase-wise breakdown of acceptance length improvements ($\Delta\tau$) and saved verification iterations per interaction turn for both the reasoning ($\tau_r$) and action ($\tau_a$) phases. After GUI-adapted fine-tuning, the action-phase gain shrinks the most, as the drafter readily memorises predefined action schemas. Consequently, the primary contribution of \SpdShort{} shifts predominantly towards the reasoning phase.

\input{tables/spd_breakdown_full}

\clearpage
\onecolumn
\section{Case Studies}
\label{app:case_study}

\subsection{\SpdLong{}}
\label{sec:case_spd}

{
\centering
\includegraphics[width=\textwidth]{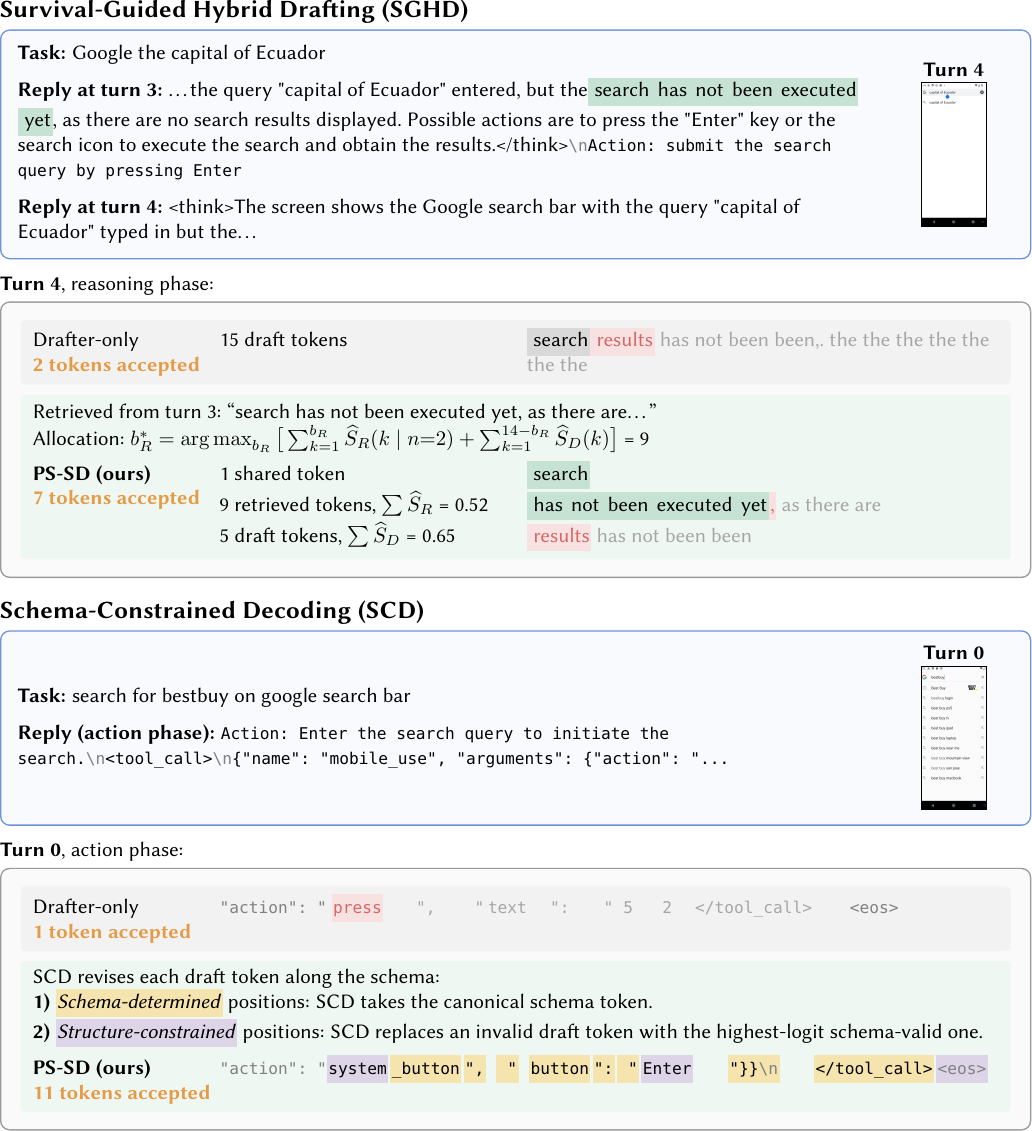}

\captionsetup{hypcap=false}
\captionof{figure}{
\textbf{PS-SD vs.\ Drafter-only.}
SGHD (top) extends drafter-only decoding with a retrieved branch via survival-guided block allocation.
SCD (bottom) revises action drafts by filling schema-determined positions with canonical tokens and replacing invalid tokens at structure-constrained positions with schema-valid ones.
}
\label{fig:case_spd}
\captionsetup{hypcap=true}
\par
}

\newpage

\subsection{\VtpLong{}}
\label{sec:case_vtp}

{
\centering
\includegraphics[width=\textwidth]{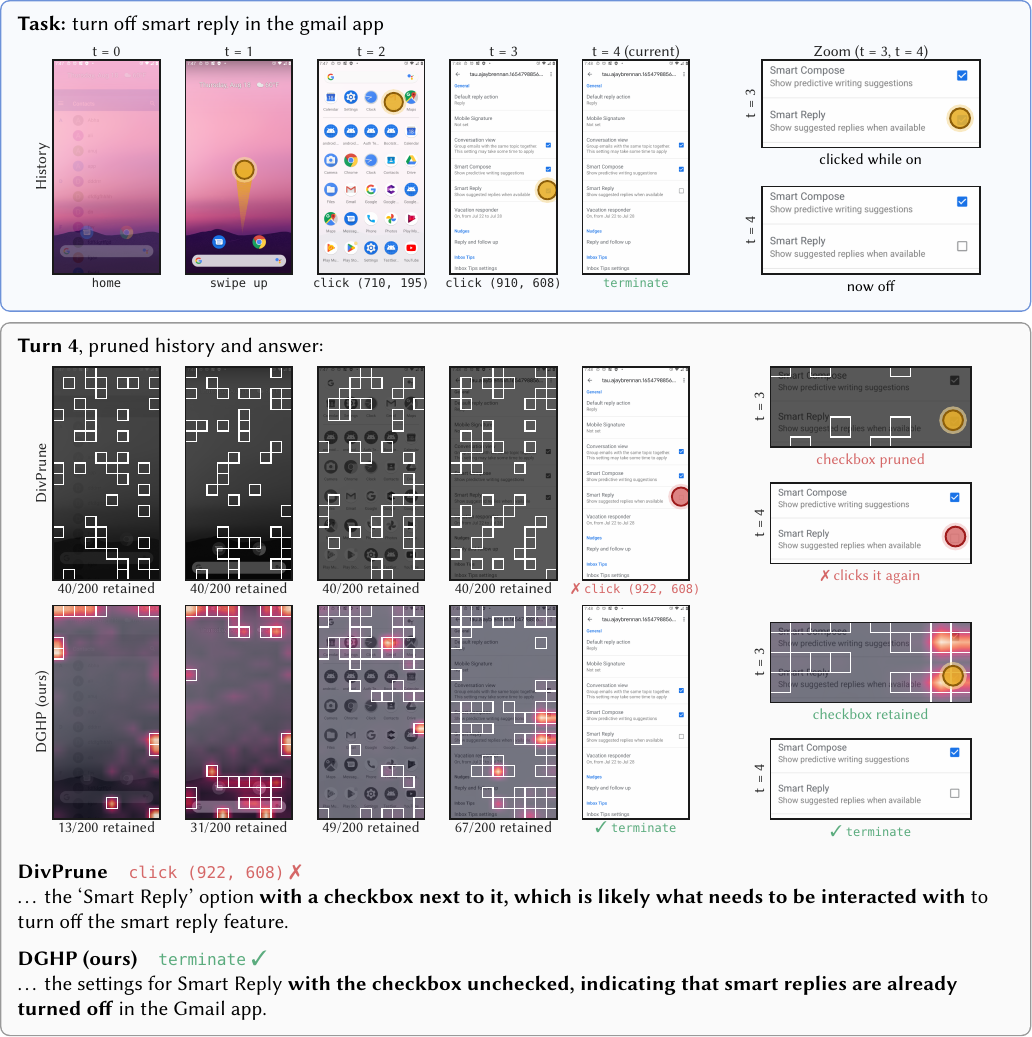}

\captionsetup{hypcap=false}
\captionof{figure}{
\textbf{\VtpLong{} vs.\ DivPrune (20\% budget).}
DivPrune applies uniform, diversity-based pruning, discarding the historical checkbox that conveys task progress. In contrast,~\VtpLong{} leverages drafter attention for task-aware historical pruning, retaining the attended checkbox at $t=3$ for correct termination.
}
\label{fig:case_vtp}
\captionsetup{hypcap=true}
\par
}

%% file: algorithms/PSSD_auxiliary.tex
{
\vspace{-5mm}

\begin{algorithm}[H]
\algrenewcommand{\algorithmicindent}{1em}
\caption{Hierarchical Suffix Retrieval}
\label{alg:retrieve}
\algtext*{EndIf}\algtext*{EndFor}\algtext*{EndWhile}\algtext*{EndFunction}
\begin{algorithmic}[1]
\Statex \textbf{Input:} Output prefix $\mathbf y$; block size $B$.
\Statex \textbf{State:} Current-episode index $\mathcal I_{\mathrm{req}}$ and index $\mathcal I_{\mathrm{corp}}$ of reasoning tokens from previous episodes.
\Statex \textbf{Parameters:} Minimum and maximum match lengths $n_{\min}=2$ and $n_{\max}=6$.
\Statex \textbf{Output:} Retrieved tokens $\mathbf r$ ($|\mathbf r|\le B-1$) and matched suffix length $n$.
\Function{Retrieve}{$\mathbf y,B$}
  \For{$\mathcal I\in(\mathcal I_{\mathrm{req}},\mathcal I_{\mathrm{corp}})$} \Comment{Current episode first}
    \For{$n=\min\{n_{\max},|\mathbf y|\}$ \textbf{down to} $n_{\min}$}
      \State $\mathbf u\gets$ last $n$ tokens of $\mathbf y$
      \If{$\mathbf u$ has a nonempty continuation in $\mathcal I$}
        \State $\mathbf r\gets$ latest such continuation
        \State \Return $(\mathbf r_{1:\min\{B-1,|\mathbf r|\}},n)$
      \EndIf
    \EndFor
  \EndFor
  \State \Return $(\varepsilon,0)$ \Comment{Empty sequence}
\EndFunction
\end{algorithmic}
\end{algorithm}
}

{
\begin{algorithm}[H]
\algrenewcommand{\algorithmicindent}{1em}
\caption{Survival-Guided Block Allocation}
\label{alg:allocate}
\algtext*{EndIf}\algtext*{EndFor}\algtext*{EndWhile}\algtext*{EndFunction}
\begin{algorithmic}[1]
\Statex \textbf{Input:} Draft tokens $\mathbf d$; draft logits $\mathbf z$; retrieved tokens $\mathbf r$; matched suffix length $n$; block size $B$.
\Statex \textbf{State:} Acceptance and verification counts $(H_{n,i}, N_{n,i})$.
\Statex \textbf{Parameters:} Prior acceptance probability $p_0=0.5$ and prior weight $w=1$.
\Statex \textbf{Output:} Candidate tree $\mathcal T$, shared prefix length $b_{\mathrm{shared}}$, and allocation $b_R^*$ for the retrieved branch.
\Function{Allocate}{$\mathbf d,\mathbf z,\mathbf r,n,B$}
  \State $b_{\mathrm{shared}}\gets$ length of the longest shared prefix of $\mathbf d$ and $\mathbf r$
  \If{$b_{\mathrm{shared}}=|\mathbf r|$} \Comment{Draft branch only}
    \State \Return $(\mathbf d,b_{\mathrm{shared}},0)$
  \EndIf
  \State $K\gets B-1-b_{\mathrm{shared}}$, \quad $L\gets\min\{K,|\mathbf r|-b_{\mathrm{shared}}\}$
  \State $c_i\gets[\operatorname{softmax}(\mathbf z_{b_{\mathrm{shared}}+i})]_{d_{b_{\mathrm{shared}}+i}}$, \quad $1\le i\le K$
  \State $\widehat S_D(k)\gets\prod_{i=1}^{k}c_i$, \quad $1\le k\le K$
  \State $\widehat S_R(k\mid n)\gets\prod_{i=1}^{k}\frac{H_{n,i}+wp_0}{N_{n,i}+w}$, \quad $1\le k\le L$
  \State Choose $b_R^*\in\{0,\ldots,L\}$ maximising the objective in~\cref{eq:hybrid_allocation}
  \State $b_D^*\gets K-b_R^*$
  \State $\mathcal T\gets\operatorname{Trie}\!\bigl(\{\mathbf d_{1:b_{\mathrm{shared}}+b_D^*},\,\mathbf r_{1:b_{\mathrm{shared}}+b_R^*}\}\bigr)$
  \State \Return $(\mathcal T,b_{\mathrm{shared}},b_R^*)$
\EndFunction
\end{algorithmic}
\end{algorithm}
}

{
\begin{algorithm}[H]
\algrenewcommand{\algorithmicindent}{1em}
\caption{Online Retrieval Acceptance Update}
\label{alg:update_acceptance}
\algtext*{EndIf}\algtext*{EndFor}\algtext*{EndWhile}\algtext*{EndFunction}
\begin{algorithmic}[1]
\Statex \textbf{Input:} Newly committed tokens $\mathbf a$; retrieved tokens $\mathbf r$; matched suffix length $n$; shared prefix length $b_{\mathrm{shared}}$; allocation $b_R^*$ for the retrieved branch.
\Statex \textbf{State:} Acceptance counts $H_{n,i}$ and verification counts $N_{n,i}$ (initially zero).
\Function{UpdateAcceptance}{$\mathbf a,\mathbf r,n,b_{\mathrm{shared}},b_R^*$}
  \For{$i=1$ \textbf{to} $\min\{b_R^*,|\mathbf a|-b_{\mathrm{shared}}\}$}
    \State $N_{n,i}\gets N_{n,i}+1$
    \If{$a_{b_{\mathrm{shared}}+i}\neq r_{b_{\mathrm{shared}}+i}$}
      \State \textbf{break}
    \EndIf
    \State $H_{n,i}\gets H_{n,i}+1$
  \EndFor
\EndFunction
\end{algorithmic}
\end{algorithm}
}

\newpage

{
\begin{algorithm}[H]
\algrenewcommand{\algorithmicindent}{1em}
\caption{Drafter-Guided Visual Token Pruning}
\label{alg:vtp}
\algtext*{EndIf}\algtext*{EndFor}\algtext*{EndWhile}\algtext*{EndFunction}
\begin{algorithmic}[1]
\Statex \textbf{Input:} Historical visual tokens $\{V_t\}_{t=1}^{T}$ in chronological order; current visual tokens $V_{\mathrm{cur}}$.
\Statex \textbf{State:} Drafter attention scores $\mathbf s_t$ from turn $t$.
\Statex \textbf{Parameters:} History retention ratio $b$ and temporal weighting parameter $\gamma\in(0,1]$.
\Statex \textbf{Output:} Retained historical visual tokens $\{\widetilde V_t\}_{t=1}^{T}$ and current visual tokens $V_{\mathrm{cur}}$.
\Function{PruneHistory}{$\{V_t\}_{t=1}^{T},V_{\mathrm{cur}}$}
  \If{$T=1$}
    \State $b_1\gets b$
  \Else
    \State $w_t\gets\gamma+(1-\gamma)\frac{t-1}{T-1}$, $t=1,\ldots,T$ \Comment{\cref{eq:temporal_weight}}
    \State $b_t\gets b\cdot T\,w_t/\sum_{s=1}^{T}w_s$, $t=1,\ldots,T$
  \EndIf
  \For{$t=1,\ldots,T$}
    \State $k_t\gets\lceil b_t|V_t|\rceil$
    \State $J_t\gets$ indices of the $k_t$ highest scores in $\mathbf s_t$
    \State $\widetilde V_t\gets$ tokens of $V_t$ at $J_t$ in their original order
  \EndFor
  \State \Return $(\{\widetilde V_t\}_{t=1}^{T},V_{\mathrm{cur}})$
\EndFunction
\end{algorithmic}
\end{algorithm}
}

%% file: tables/spd_breakdown_full.tex
\begin{table*}[t]
\centering
\small
\setlength{\tabcolsep}{2pt}
\renewcommand{\arraystretch}{0.95}
\setlength{\aboverulesep}{0.5ex}
\setlength{\belowrulesep}{0.5ex}
\caption{
\textbf{Phase-wise breakdown of \SpdShort{} on GUI-Owl-1.5-8B.}
}
\label{tab:phase_breakdown_owl}
\begin{tabularx}{\textwidth}{
    @{}
    l
    @{\hspace{0.8em}}
    *{12}{>{\centering\arraybackslash}X}
    @{}
}
\toprule
& \multicolumn{6}{c}{\makebox[0pt][c]{\textbf{DFlash}}}
& \multicolumn{6}{c}{\makebox[0pt][c]{\textbf{Domino}}} \\
\cmidrule(l{0pt}r{3pt}){2-7}
\cmidrule(l{3pt}r{0pt}){8-13}
& \multicolumn{2}{c}{\makebox[0pt][c]{GUI-Ody}}
& \multicolumn{2}{c}{\makebox[0pt][c]{AITW}}
& \multicolumn{2}{c}{\makebox[0pt][c]{M2Web}}
& \multicolumn{2}{c}{\makebox[0pt][c]{GUI-Ody}}
& \multicolumn{2}{c}{\makebox[0pt][c]{AITW}}
& \multicolumn{2}{c}{\makebox[0pt][c]{M2Web}} \\
\cmidrule(l{0pt}r{3pt}){2-3}
\cmidrule(l{3pt}r{3pt}){4-5}
\cmidrule(l{3pt}r{3pt}){6-7}
\cmidrule(l{3pt}r{3pt}){8-9}
\cmidrule(l{3pt}r{3pt}){10-11}
\cmidrule(l{3pt}r{0pt}){12-13}
\textbf{Variant}
& $\tau_r$ & $\tau_a$ & $\tau_r$ & $\tau_a$ & $\tau_r$ & $\tau_a$
& $\tau_r$ & $\tau_a$ & $\tau_r$ & $\tau_a$ & $\tau_r$ & $\tau_a$ \\[-0.3ex]
\midrule
General (G)
& 2.37 & 2.53 & 2.26 & 3.10 & 2.43 & 3.08
& 2.48 & 2.55 & 2.50 & 3.03 & 2.64 & 3.16 \\
\quad +\textit{\SpdShort}
& \textbf{3.90} & \textbf{4.94} & \textbf{3.04} & \textbf{4.69} & \textbf{2.74} & \textbf{4.24}
& \textbf{4.02} & \textbf{5.04} & \textbf{3.22} & \textbf{4.86} & \textbf{2.94} & \textbf{4.31} \\
$\Delta\tau$ (\%)
& \deltaval{+}{64.6} & \deltaval{+}{95.3} & \deltaval{+}{34.5} & \deltaval{+}{51.3} & \deltaval{+}{12.8} & \deltaval{+}{37.7}
& \deltaval{+}{62.1} & \deltaval{+}{97.6} & \deltaval{+}{28.8} & \deltaval{+}{60.4} & \deltaval{+}{11.4} & \deltaval{+}{36.4} \\
Saved steps
& 20.0 & 9.1 & 6.8 & 5.2 & 5.2 & 5.0
& 18.7 & 9.1 & 5.4 & 6.0 & 4.3 & 4.7 \\[-0.3ex]
\midrule
Agent (A)
& 3.85 & 6.63 & 2.49 & 5.66 & 2.74 & 4.83
& 4.49 & 6.95 & 2.79 & 6.10 & 3.05 & 5.06 \\
\quad +\textit{\SpdShort}
& \textbf{4.82} & \textbf{6.79} & \textbf{3.25} & \textbf{5.87} & \textbf{2.97} & \textbf{5.06}
& \textbf{5.35} & \textbf{6.96} & \textbf{3.47} & \textbf{6.13} & \textbf{3.26} & \textbf{5.30} \\
$\Delta\tau$ (\%)
& \deltaval{+}{25.2} & \deltaval{+}{2.41} & \deltaval{+}{30.5} & \deltaval{+}{3.71} & \deltaval{+}{8.39} & \deltaval{+}{4.76}
& \deltaval{+}{19.2} & \deltaval{+}{0.14} & \deltaval{+}{24.4} & \deltaval{+}{0.49} & \deltaval{+}{6.89} & \deltaval{+}{4.74} \\
Saved steps
& 6.3 & 0.2 & 5.6 & 0.3 & 3.2 & 0.5
& 4.3 & 0.0 & 4.2 & 0.0 & 2.4 & 0.5 \\[-0.3ex]
\midrule
Avg. tokens
& 121 & 47 & 60 & 48 & 112 & 56
& 121 & 47 & 60 & 48 & 112 & 56 \\[-0.3ex]
\bottomrule
\end{tabularx}
\end{table*}

\begin{table*}[t]
\centering
\small
\setlength{\tabcolsep}{2pt}
\renewcommand{\arraystretch}{0.95}
\setlength{\aboverulesep}{0.5ex}
\setlength{\belowrulesep}{0.5ex}
\caption{
\textbf{Phase-wise breakdown of \SpdShort{} on MAI-UI-8B.}
}
\label{tab:phase_breakdown_mai}
\begin{tabularx}{\textwidth}{
    @{}
    l
    @{\hspace{0.8em}}
    *{12}{>{\centering\arraybackslash}X}
    @{}
}
\toprule
& \multicolumn{6}{c}{\makebox[0pt][c]{\textbf{DFlash}}}
& \multicolumn{6}{c}{\makebox[0pt][c]{\textbf{Domino}}} \\
\cmidrule(l{0pt}r{3pt}){2-7}
\cmidrule(l{3pt}r{0pt}){8-13}
& \multicolumn{2}{c}{\makebox[0pt][c]{GUI-Ody}}
& \multicolumn{2}{c}{\makebox[0pt][c]{AITW}}
& \multicolumn{2}{c}{\makebox[0pt][c]{M2Web}}
& \multicolumn{2}{c}{\makebox[0pt][c]{GUI-Ody}}
& \multicolumn{2}{c}{\makebox[0pt][c]{AITW}}
& \multicolumn{2}{c}{\makebox[0pt][c]{M2Web}} \\
\cmidrule(l{0pt}r{3pt}){2-3}
\cmidrule(l{3pt}r{3pt}){4-5}
\cmidrule(l{3pt}r{3pt}){6-7}
\cmidrule(l{3pt}r{3pt}){8-9}
\cmidrule(l{3pt}r{3pt}){10-11}
\cmidrule(l{3pt}r{0pt}){12-13}
\textbf{Variant}
& $\tau_r$ & $\tau_a$ & $\tau_r$ & $\tau_a$ & $\tau_r$ & $\tau_a$
& $\tau_r$ & $\tau_a$ & $\tau_r$ & $\tau_a$ & $\tau_r$ & $\tau_a$ \\[-0.3ex]
\midrule
General (G)
& 2.09 & 1.65 & 2.51 & 2.25 & 2.67 & 2.05
& 2.17 & 1.46 & 2.68 & 2.22 & 2.86 & 2.13 \\
\quad +\textit{\SpdShort}
& \textbf{3.37} & \textbf{4.44} & \textbf{3.88} & \textbf{4.65} & \textbf{3.38} & \textbf{3.94}
& \textbf{3.36} & \textbf{4.16} & \textbf{3.93} & \textbf{4.70} & \textbf{3.48} & \textbf{4.05} \\
$\Delta\tau$ (\%)
& \deltaval{+}{61.2} & \deltaval{+}{169} & \deltaval{+}{54.6} & \deltaval{+}{107} & \deltaval{+}{26.6} & \deltaval{+}{92.2}
& \deltaval{+}{54.8} & \deltaval{+}{185} & \deltaval{+}{46.6} & \deltaval{+}{112} & \deltaval{+}{21.7} & \deltaval{+}{90.1} \\
Saved steps
& 8.7 & 10.7 & 7.5 & 6.2 & 4.9 & 6.6
& 7.8 & 12.4 & 6.3 & 6.4 & 3.9 & 6.2 \\[-0.3ex]
\midrule
Agent (A)
& 4.21 & 5.17 & 4.44 & 5.11 & 4.19 & 4.32
& 4.88 & 5.23 & 5.25 & 5.28 & 4.89 & 4.33 \\
\quad +\textit{\SpdShort}
& \textbf{4.87} & \textbf{5.46} & \textbf{5.26} & \textbf{5.41} & \textbf{4.60} & \textbf{4.49}
& \textbf{5.44} & \textbf{5.32} & \textbf{5.87} & \textbf{5.39} & \textbf{5.22} & \textbf{4.37} \\
$\Delta\tau$ (\%)
& \deltaval{+}{15.7} & \deltaval{+}{5.61} & \deltaval{+}{18.5} & \deltaval{+}{5.87} & \deltaval{+}{9.79} & \deltaval{+}{3.94}
& \deltaval{+}{11.5} & \deltaval{+}{1.72} & \deltaval{+}{11.8} & \deltaval{+}{2.08} & \deltaval{+}{6.75} & \deltaval{+}{0.92} \\
Saved steps
& 1.5 & 0.3 & 1.9 & 0.3 & 1.3 & 0.2
& 1.0 & 0.1 & 1.1 & 0.1 & 0.8 & 0.1 \\[-0.3ex]
\midrule
Avg. tokens
& 48 & 28 & 53 & 27 & 62 & 28
& 48 & 28 & 53 & 27 & 62 & 28 \\[-0.3ex]
\bottomrule
\end{tabularx}
\end{table*}

\begin{table*}[t]
\centering
\small
\setlength{\tabcolsep}{2pt}
\renewcommand{\arraystretch}{0.95}
\setlength{\aboverulesep}{0.5ex}
\setlength{\belowrulesep}{0.5ex}
\caption{
\textbf{Phase-wise breakdown of \SpdShort{} on Qwen3-VL-8B.}
}
\label{tab:phase_breakdown_qwen}
\begin{tabularx}{\textwidth}{
    @{}
    l
    @{\hspace{0.8em}}
    *{12}{>{\centering\arraybackslash}X}
    @{}
}
\toprule
& \multicolumn{6}{c}{\makebox[0pt][c]{\textbf{DFlash}}}
& \multicolumn{6}{c}{\makebox[0pt][c]{\textbf{Domino}}} \\
\cmidrule(l{0pt}r{3pt}){2-7}
\cmidrule(l{3pt}r{0pt}){8-13}
& \multicolumn{2}{c}{\makebox[0pt][c]{GUI-Ody}}
& \multicolumn{2}{c}{\makebox[0pt][c]{AITW}}
& \multicolumn{2}{c}{\makebox[0pt][c]{M2Web}}
& \multicolumn{2}{c}{\makebox[0pt][c]{GUI-Ody}}
& \multicolumn{2}{c}{\makebox[0pt][c]{AITW}}
& \multicolumn{2}{c}{\makebox[0pt][c]{M2Web}} \\
\cmidrule(l{0pt}r{3pt}){2-3}
\cmidrule(l{3pt}r{3pt}){4-5}
\cmidrule(l{3pt}r{3pt}){6-7}
\cmidrule(l{3pt}r{3pt}){8-9}
\cmidrule(l{3pt}r{3pt}){10-11}
\cmidrule(l{3pt}r{0pt}){12-13}
\textbf{Variant}
& $\tau_r$ & $\tau_a$ & $\tau_r$ & $\tau_a$ & $\tau_r$ & $\tau_a$
& $\tau_r$ & $\tau_a$ & $\tau_r$ & $\tau_a$ & $\tau_r$ & $\tau_a$ \\[-0.3ex]
\midrule
General (G)
& 2.00 & 2.28 & 2.46 & 3.64 & 2.31 & 3.27
& 2.40 & 2.70 & 2.80 & 3.69 & 2.52 & 3.37 \\
\quad +\textit{\SpdShort}
& \textbf{2.84} & \textbf{4.59} & \textbf{3.05} & \textbf{5.33} & \textbf{2.64} & \textbf{4.50}
& \textbf{3.16} & \textbf{4.91} & \textbf{3.33} & \textbf{5.68} & \textbf{2.83} & \textbf{5.04} \\
$\Delta\tau$ (\%)
& \deltaval{+}{42.0} & \deltaval{+}{101} & \deltaval{+}{24.0} & \deltaval{+}{46.4} & \deltaval{+}{14.3} & \deltaval{+}{37.6}
& \deltaval{+}{31.7} & \deltaval{+}{81.9} & \deltaval{+}{18.9} & \deltaval{+}{53.9} & \deltaval{+}{12.3} & \deltaval{+}{49.6} \\
Saved steps
& 17.6 & 10.6 & 9.1 & 4.0 & 14.2 & 4.6
& 11.9 & 8.0 & 6.6 & 4.4 & 11.4 & 5.4 \\[-0.3ex]
\midrule
Agent (A)
& 3.37 & 8.04 & 3.02 & 7.36 & 2.49 & 5.52
& 3.97 & 8.47 & 3.48 & \textbf{7.79} & 2.70 & 5.64 \\
\quad +\textit{\SpdShort}
& \textbf{3.85} & \textbf{8.15} & \textbf{3.46} & \textbf{7.46} & \textbf{2.80} & \textbf{6.09}
& \textbf{4.37} & 8.47 & \textbf{3.86} & 7.78 & \textbf{2.98} & \textbf{6.36} \\
$\Delta\tau$ (\%)
& \deltaval{+}{14.2} & \deltaval{+}{1.37} & \deltaval{+}{14.6} & \deltaval{+}{1.36} & \deltaval{+}{12.4} & \deltaval{+}{10.3}
& \deltaval{+}{10.1} & 0.00 & \deltaval{+}{10.9} & \deltaval{-}{0.13} & \deltaval{+}{10.4} & \deltaval{+}{12.8} \\
Saved steps
& 4.4 & 0.1 & 4.9 & 0.1 & 11.6 & 0.9
& 2.7 & 0.0 & 3.3 & 0.0 & 9.1 & 1.1 \\[-0.3ex]
\midrule
Avg. tokens
& 119 & 48 & 116 & 46 & 262 & 55
& 119 & 48 & 116 & 46 & 262 & 55 \\[-0.3ex]
\bottomrule
\end{tabularx}
\end{table*}